\documentclass[sigconf]{acmart}

\usepackage{booktabs}
\usepackage{multirow}
\usepackage{graphicx}
\usepackage[most]{tcolorbox}
\usepackage{stfloats}
\usepackage{enumitem}
\usepackage{url}
\AtBeginDocument{%
  }

\setcopyright{acmlicensed}
\copyrightyear{2026}
\acmYear{2026}
\acmDOI{XXXXXXX.XXXXXXX}
\acmConference[KDD '26]{Make sure to enter the correct
  conference title from your rights confirmation email}{August 9--13,
  2026}{Jeju, Korea}
\acmISBN{978-1-4503-XXXX-X/2018/06}

\begin{document}

\title{Beyond One-Size-Fits-All: Adaptive Subgraph Denoising for Zero-Shot Graph Learning with Large Language Models}

\author{Fengzhi Li}
\affiliation{%
  \institution{JIUTIAN Research}
  \city{Beijing}
  \country{China}
}
\email{lifengzhi@cmjt.chinamobile.com}

\author{Liang Zhang}
\authornote{Corresponding Author}
\affiliation{%
  \institution{The Hong Kong University of Science and Technology (Guangzhou)}
  \city{Guangzhou}
  \country{China}
}
\email{liangzhang@hkust-gz.edu.cn}

\author{Yuan Zuo}
\affiliation{%
  \institution{MIIT Key Laboratory of Data and Decision Intelligence}
  \institution{Beihang University}
  \city{Beijing}
  \country{China}
}
\email{zuoyuan@buaa.edu.cn}

\author{Ruiqing Zhao}
\affiliation{%
  \institution{Beihang University}
  \city{Beijing}
  \country{China}
}
\email{ruiqingzhao@buaa.edu.cn}

\author{Yansong Liu}
\affiliation{%
 \institution{Beihang University}
 \city{Beijing}
 \country{China}
}
\email{liuyansong@buaa.edu.cn}

\author{Yunfei Ma}
\affiliation{%
  \institution{JIUTIAN Research}
  \city{Beijing}
  \country{China}
}
\email{mayunfei@cmjt.chinamobile.com}

\author{Fanyu Meng}
\affiliation{%
  \institution{JIUTIAN Research}
  \city{Beijing}
  \country{China}
}
\email{mengfanyu@cmjt.chinamobile.com}

\author{Junlan Feng}
\affiliation{%
  \institution{JIUTIAN Research}
  \city{Beijing}
  \country{China}
}
\email{fengjunlan@cmjt.chinamobile.com}

\renewcommand{\shortauthors}{Fengzhi Li et al.}

\begin{abstract}
  Graph-based tasks in the zero-shot setting remain a significant challenge due to data scarcity and the inability of traditional Graph Neural Networks (GNNs) to generalize to unseen domains or label spaces. While recent advancements have transitioned toward leveraging Large Language Models (LLMs) as predictors to enhance GNNs, these methods often suffer from cross-modal alignment issues. A recent paradigm (\textit{i.e.}, Graph-R1) overcomes the aforementioned architectural dependencies by adopting a purely text-based format and utilizing LLM-based graph reasoning, showing improved zero-shot generalization. However, it employs a task-agnostic, one-size-fits-all subgraph extraction strategy, which inevitably introduces significant structural noise—irrelevant neighbors and edges—that distorts the LLMs' receptive field and leads to suboptimal predictions. To address this limitation, we introduce \textbf{GraphSSR}, a novel framework designed for adaptive subgraph extraction and denoising in zero-shot LLM-based graph reasoning. Specifically, we propose the \textbf{SSR} pipeline, which dynamically tailors subgraph extraction to specific contexts through a ``Sample-Select-Reason'' process, enabling the model to autonomously filter out task-irrelevant neighbors and overcome the one-size-fits-all issue. To internalize this capability, we develop \textbf{SSR-SFT}, a data synthesis strategy that generates high-quality SSR-style graph reasoning traces for supervised fine-tuning of LLMs. Furthermore, we propose \textbf{SSR-RL}, a two-stage reinforcement learning framework that explicitly regulates sampling and selection operations within the proposed SSR pipeline designed for adaptive subgraph denoising. By incorporating Authenticity-Reinforced and Denoising-Reinforced RL, we guide the model to achieve accurate predictions using parsimonious, denoised subgraphs for reasoning. Extensive experiments across multiple benchmarks demonstrate that GraphSSR significantly outperforms state-of-the-art methods, highlighting the necessity of adaptive subgraph denoising for zero-shot graph reasoning. Our code is publicly available at \url{ https://github.com/mysteriouslfz/GraphSSR}.
\end{abstract}
 %



\received{9 February 2026}
\received[revised]{17 April 2026}
\received[accepted]{16 May 2026}

\maketitle

\section{Introduction}\label{sec:intro}

Graph-based tasks, such as node classification and link prediction, are ubiquitous across various real-world domains, from social network analysis and bioinformatics to advanced recommendation systems \cite{zhang2022improving, guo2025fgdgnn, yi2022graph, zhang2025graph, gao2023survey, el2025novel}. However, traditional methods for these tasks often rely on sufficient labeled data for model training, which can be particularly challenging in many real-world scenarios, especially when dealing with new or unseen domains. This data scarcity issue has recently sparked growing research interest in the zero-shot learning setting for graphs \cite{zhao2023gimlet, chen2024review, zhao2025dynamic}, where models are able to perform tasks in previously unseen label spaces or domains without the need for task-specific supervision.

Graph neural networks (GNNs) are the leading methods designed specifically for graph-based tasks, significantly advancing the field by enabling models to capture intricate structural dependencies. However, while conventional GNN-based methods perform well with ample labeled data, their generalization ability deteriorates sharply under distribution shifts or in unseen label spaces, \textit{i.e.}, in the so-called zero-shot learning setting. In this context, Large Language Models (LLMs) offer a promising complementary alternative. 

Existing methods generally fall into two strategic paradigms: LLM-as-enhancer and LLM-as-predictor. The former, including methods such as OFA~\cite{liu2023one}, ZeroG~\cite{li2024zerog}, and DAS~\cite{thapaliya2025semantic}, uses LLMs to refine node features or generate auxiliary signals (\textit{e.g.}, synthetic labels or node descriptions). However, they remain tied to GNN backbones for prediction, which still require retraining for each task based on labeled data, limiting their transferability to unseen domains or labels. On the other hand, LLM-as-predictor methods, such as GraphGPT~\cite{tang2024graphgpt} and GOFA~\cite{kong2024gofa}, directly leverage LLMs as inference engines, where structural signals from a frozen GNN serve as input through cross-modal projection or token insertions. While this paradigm improves zero-shot accuracy to some extent by leveraging the generalization of LLM predictions, the separation between the GNN and LLM components, each operating in distinct modalities, requires complex alignment pretraining. Such alignment may not hold in new domains or label spaces, severely hindering true zero-shot generalization.

To overcome the architectural dependencies in previous LLM-based modeling paradigms, a recent work, Graph-R1~\cite{wu2025graph}, rethinks zero-shot graph reasoning in a purely text-based format. It demonstrates that LLM-based reasoning over graphs, empowered by instruction tuning and reinforcement learning, can significantly enhance zero-shot generalization. Taking zero-shot node classification as an example, Graph-R1 transforms a target node and its associated $h$-hop neighbor subgraph, along with task descriptions, into a text sequence. It then prompts an LLM to generate reasoning chains for prediction via a transparent reason-then-predict pipeline.

Although successful, Graph-R1 assumes a uniform and task-agnostic subgraph extraction setting in its reasoning design: for any node and task (\textit{e.g.}, label space or domain), the same $h$-hop local subgraph extraction strategy is applied for LLM reasoning. However, we argue that such a one-size-fits-all approach is inherently suboptimal, as real-world graphs often contain prevalent structural noise~\cite{ju2025survey}. In Graph-R1, irrelevant nodes and edges, which do not contribute to the task at hand, are inevitably included. These task-irrelevant components introduce noise, distorting the model's receptive field and interfering with its reasoning process, ultimately leading to suboptimal or even erroneous predictions. 

\begin{figure}[t!]
    \centering
    \includegraphics[width=1.0\linewidth]{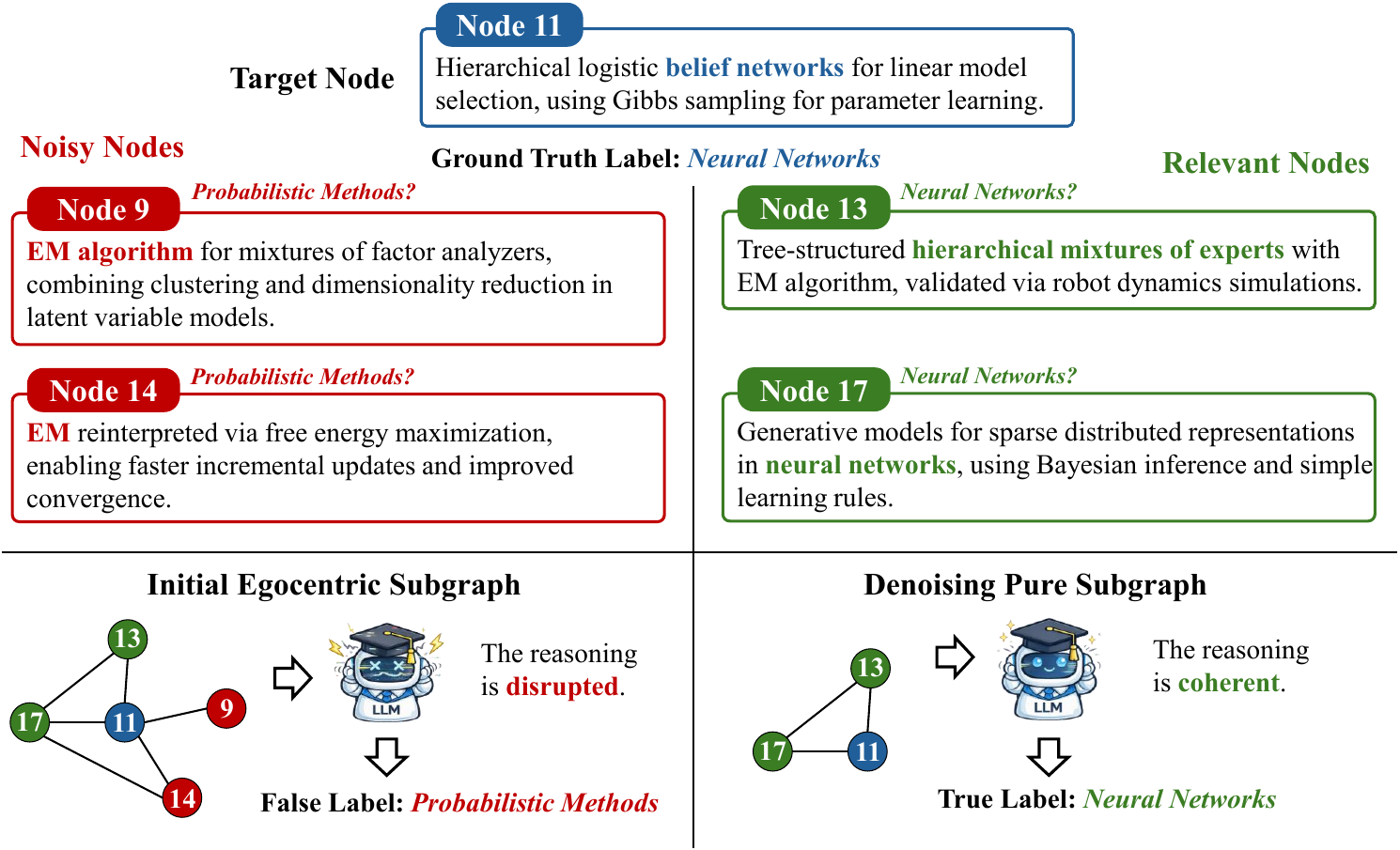} 
    \Description{A case of structural noise in the subgraph misleading the model's reasoning from the Cora dataset.}
    \caption{A case of structural noise in the subgraph misleading the model's reasoning from the Cora dataset.}
    \label{fig:intro_case}
    \vspace{-10pt}
\end{figure}

To illustrate this challenge, consider a specific instance from the Cora~\cite{wen2023augmenting} dataset. As shown in Figure~\ref{fig:intro_case}, the target node describes: ``\textit{Hierarchical logistic belief networks for linear model selection, using Gibbs sampling for parameter learning}'', belonging to the ground-truth category ``\textbf{Neural Networks}''. Among its $h$-hop local subgraph, some neighboring nodes are semantically aligned, such as node 13 (discussing hierarchical mixtures of experts) and node 17 (discussing models of neural networks). However, the subgraph also contains noisy nodes like node 9 and node 14 (both discussing different variants of the EM algorithm). While these neighbors are topologically connected, their semantic focus shifts toward a different category ``\textbf{Probabilistic Methods}''. Without an explicit mechanism to prune these irrelevant nodes, the LLM is prone to aggregating misleading ``Probabilistic Methods'' context, leading to a prediction biased by structural noise and resulting in misclassification. Conversely, if the model can autonomously filter out these noisy neighbors and focus solely on relevant ``Neural Networks'' structures, the quality of the reasoning process will be enhanced.

To address the challenges of task-agnostic subgraphs and structural noise, we propose a novel framework, \textbf{GraphSSR}, for subgraph denoising in zero-shot LLM-based graph reasoning. Central to our method is the \textbf{``Sample-Select-Reason'' (SSR)} pipeline, a meticulously designed graph reasoning process that dynamically tailors subgraph extraction to both the specific node (\textit{e.g.}, zero-shot node classification) and task conditions, avoiding the one-size-fits-all limitation. Specifically, in the first stage, the model \textbf{Samples} a diverse set of candidate subgraphs, exploring various structural and semantic perspectives of the target node. Subsequently, the model performs a dynamic denoising process by evaluating the quality of each candidate subgraph, discarding those contaminated by irrelevant nodes or edges, and \textbf{Selecting} only the most ``pure'' structural context. This approach explicitly mitigates the impact of structural noise. Finally, the model executes high-fidelity \textbf{Reasoning} on the filtered subgraph, leveraging its task-specific context to generate accurate predictions. To empower the LLM model with such specialized graph reasoning capabilities, we introduce a novel data synthesis strategy, \textbf{SSR-SFT}, for Supervised Fine-Tuning (SFT). By leveraging the open-source Graph-R1~\cite{wu2025graph} dataset and advanced teacher models, we curate high-quality graph reasoning traces that explicitly guide the model in executing the SSR pipeline.

Building on the SSR pipeline, we introduce \textbf{SSR-RL}, a two-stage reinforcement learning framework to further enhance the model's subgraph denoising and reasoning abilities using Group Relative Policy Optimization (GRPO)~\cite{shao2024deepseekmath}. Unlike traditional outcome-based RL, which focuses solely on final answer correctness, we design intermediate rewards to directly guide the subgraph denoising steps, \textit{i.e.}, sampling and selecting operations within the SSR pipeline, tailored for zero-shot graph reasoning. Specifically, in the first stage, we propose \textbf{Authenticity-Reinforced RLVR}, which incorporates sampling authenticity and selection consistency rewards within the standard RL from Verifiable Rewards (RLVR) framework. This stage helps reduce hallucinations during subgraph sampling and selecting while solidifying the model's foundational graph reasoning capabilities. In the second stage, we further propose \textbf{Denoising-Reinforced RLVR}, where a subgraph size-based reward mechanism guides the subgraph denoising process by assigning additional rewards to correct reasoning derived from more parsimonious subgraphs. This stage encourages the model to perform structural denoising and filter out irrelevant nodes and edges, effectively penalizing structural noise. Extensive experimental results demonstrate that our method significantly enhances reasoning precision in complex graph environments, achieving state-of-the-art performance across multiple zero-shot graph benchmarks. Our primary contributions are summarized as follows:
\begin{itemize}[leftmargin=*, noitemsep]
    \item \textbf{Conceptually:} For the first time, we rethink the existing LLM-based zero-shot reasoning pipeline, revealing the inherent limitations of traditional one-size-fits-all subgraph extraction strategies, and then reformulate zero-shot graph reasoning as a ``Sample-Select-Reason'' process, enabling autonomous and adaptive structural denoising.
    \item \textbf{Methodologically:} We design a new post-training framework for LLMs built upon the SSR pipeline, where SSR-SFT introduces a rigorous data synthesis and filtering strategy to create high-quality SSR-style graph reasoning demonstrations for supervised fine-tuning. This is followed by a two-stage reinforcement learning framework, \textit{i.e.}, Authenticity-Reinforced RLVR and Denoising-Reinforced RLVR, that incorporates intermediate rewards to directly guide the subgraph sampling and selecting operations, encouraging the use of more parsimonious subgraphs and reducing structural noise.
    \item \textbf{Empirically:} Extensive experiments show that our method significantly outperforms existing LLM-based graph learning methods and vanilla LLMs of the same scale across various zero-shot graph-based tasks of benchmark datasets.
\end{itemize}

\section{Methodology}

To address the inherent challenges of task-agnostic subgraphs and associated structural noise in zero-shot graph reasoning, we propose a novel framework, \textbf{GraphSSR}. It is built upon the \textbf{Sample-Select-Reason (SSR)} pipeline, which shifts subgraph extraction from a one-size-fits-all strategy to an adaptive, task-specific approach. 

\begin{figure*}[htp!]
    \centering
    \includegraphics[width=0.76\linewidth]{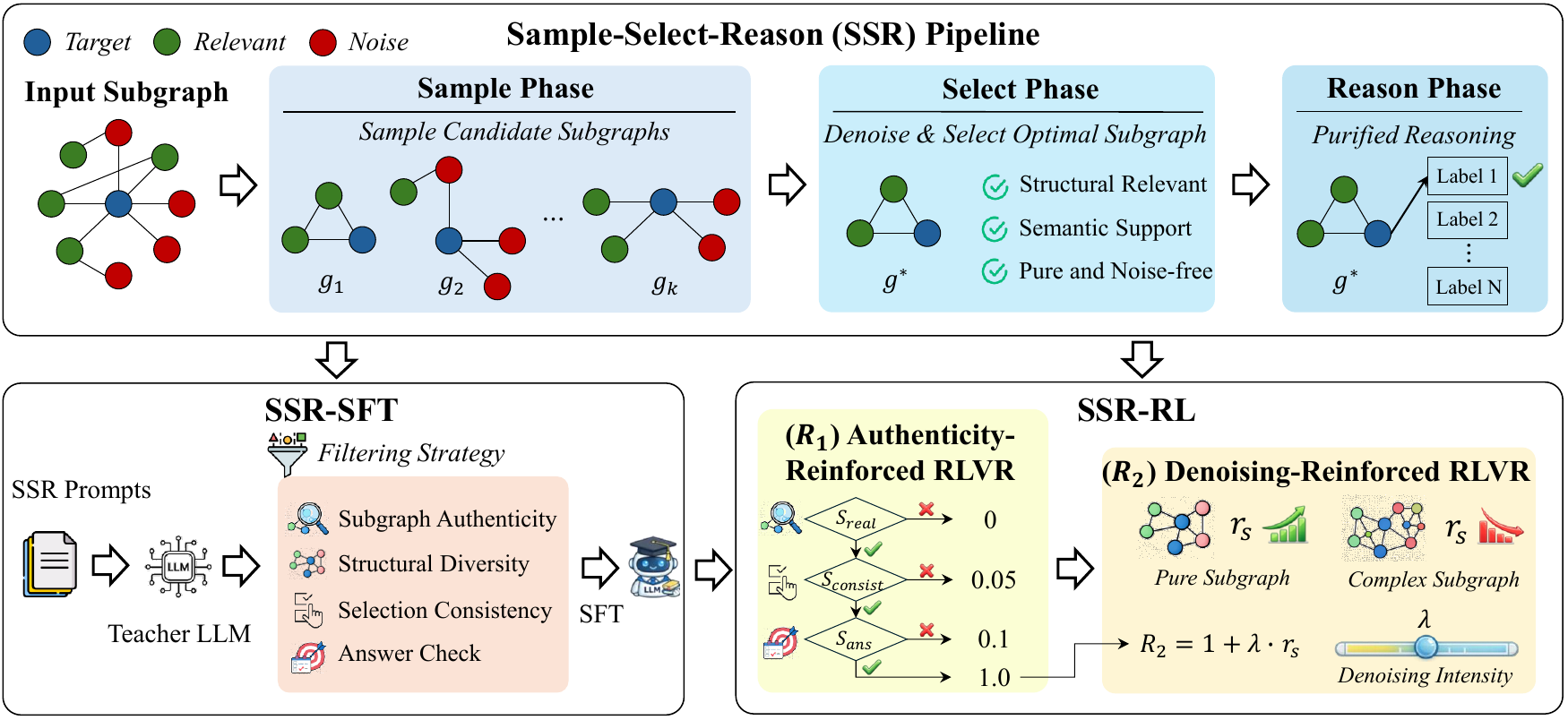} 
    \Description{Overview of our proposed \textbf{GraphSSR} framework.}
    \caption{Overview of our proposed \textbf{GraphSSR} framework.}
    \label{fig:framework}
\end{figure*}

Specifically, we reformulate graph-based tasks (\textit{e.g.}, node classification) as a conditional generation problem in which the model first executes a \textbf{Sample Phase} to explore a diverse space of candidate subgraphs, followed by a \textbf{Select Phase} to autonomously prune task-irrelevant noisy ones, and concludes with a \textbf{Reason Phase} that leverages the denoised structural context to generate answers. Within this conceptual framework, the sampling and selecting operations mitigate the impact of structural noise by explicitly performing subgraph denoising, thereby focusing on the most ``pure'' structural context. To empower the model with SSR-based graph reasoning capabilities, we introduce \textbf{SSR-SFT}, a rigorous data synthesis strategy that leverages teacher models and multi-dimensional quality filters to curate high-quality reasoning traces. To further enhance the model's subgraph denoising and reasoning abilities, we introduce \textbf{SSR-RL}, a two-stage reinforcement learning framework: the first stage strengthens reasoning and reduces hallucinations during subgraph denoising by incorporating sampling authenticity and selection consistency rewards, while the second stage introduces a subgraph size-based reward that explicitly penalizes structural noise and encourages parsimonious subgraph selection. The overall architecture of our proposed GraphSSR is illustrated in Figure~\ref{fig:framework}.

\subsection{Problem Definition}

We represent a graph as $\mathcal{G}=(\mathcal{V}, \mathcal{E}, \mathbf{A}, \mathbf{X})$, where $\mathcal{V}=\left\{v_{1}, v_{2},\ldots, v_{n}\right\}$ denotes the set of $n$ nodes and $\mathcal{E} \subseteq \mathcal{V} \times \mathcal{V}$ represents the set of edges. The structural topology is captured by the adjacency matrix $\mathbf{A} \in \{0, 1\}^{n \times n}$, with $\mathbf{A}_{ij} = 1$ if an edge exists between $v_{i}$ and $v_{j}$, and $\mathbf{A}_{ij} = 0$ otherwise. Each node $v_{i} \in \mathcal{V}$ is associated with a textual attribute $\mathbf{x}_{i} \in \mathbf{X}$, providing semantic context to the structural information. 

Given a specific zero-shot graph-based task $\tau$ (\textit{e.g.}, node classification or link classification), LLM-based graph reasoning typically reformulates the task as a conditional generation problem. This involves extracting a subgraph $\mathcal{G}_{\text{sub}} \subset \mathcal{G}$ to preserve relevant structural information, and applying a transformation function $\mathcal{T}(\cdot)$ that maps the subgraph $\mathcal{G}_{\text{sub}}$, its associated text attributes $\mathbf{X}_{\text{sub}}$, the candidate label set $\mathcal{Y}$, and the task description $\mathcal{D}$ into a natural language prompt. The graph-based task $\mathcal{\tau}$ is then defined as optimizing an LLM model that maximizes the probability of generating the correct label $y$ given the prompt, formulated as $P \left( y \mid \mathcal{T}(\mathcal{G}_{\text{sub}}, \mathbf{X}_{\text{sub}}, \mathcal{Y}, \mathcal{D}) \right)$.



\subsection{Sample-Select-Reason (SSR) Pipeline}\label{sec:method_ssr}

Given the formulation $P \left( y \mid \mathcal{T}(\mathcal{G}_{\text{sub}}, \mathbf{X}_{\text{sub}}, \mathcal{Y}, \mathcal{D}) \right)$, we identify subgraph extraction as the key to the success of zero-shot graph reasoning. Existing methods predominantly employ a one-size-fits-all subgraph extraction strategy (\textit{e.g.}, all $h$-hop neighbors). However, such a task-agnostic strategy often includes task-irrelevant nodes or edges and introduces significant structural noise, making it suboptimal. To address this issue, we reformulate the zero-shot graph reasoning task as a systematic \textit{``Sample-Select-Reason (SSR)''} pipeline designed for subgraph denoising and structural refinement to enable adaptive subgraph extraction. Specific prompt designs for the SSR pipeline are detailed in Online Appendix~\ref{app:prompt_ssr}.


\paragraph{\textbf{Sample Phase.}} Given a specific task $\tau$, our goal is to extract an optimized subgraph $\mathcal{G}_{\tau}$ suitable for graph reasoning, centered around the target node(s) associated with $\tau$ (\textit{e.g.}, node pairs for link classification). Recognizing that generating the optimal subgraph in a single shot is inherently challenging due to topological complexity, we introduce a diversity-driven subgraph group generation strategy, inspired by the group-based reasoning in the GRPO algorithm. Rather than sampling a single definitive subgraph, the model samples a group of candidates $\mathbf{S} = \{g_1, g_2, \dots, g_k\}$, analogous to GRPO's multi-rollout process. Each candidate $g_i$ represents a distinct structural and textual perspective of the local neighborhood. This exploration of the ``subgraph space'' ensures sufficient subgraph diversity, significantly increasing the likelihood that at least one candidate captures the task-relevant structures necessary for the targeted task $\tau$.

\paragraph{\textbf{Select Phase.}} We implement a denoising mechanism where the model is designed to evaluate the sampled candidates and select the most suitable subgraph for the specific downstream task $\tau$. By analyzing the structural and textual properties of each $g_i \in \mathbf{S}$, the model autonomously identifies and retains the optimal subgraph $g^*$ while discarding those containing noisy information. The Sample and Select phases work together to perform adaptive subgraph denoising, effectively pruning irrelevant graph features prior to the final reasoning step.

\paragraph{\textbf{Reason Phase.}} The refined subgraph $g^*$ serves as the context for the downstream task $\tau$. Having been denoised through the Sample and Select phases, $g^*$ provides purified structural and textual signals, allowing the LLM to avoid interference from noise and improving the performance of zero-shot graph reasoning. 


\subsection{SSR-SFT}\label{sec:method_sft}

To enable the model to execute the SSR pipeline, we construct a high-quality Supervised Fine-Tuning (SFT) dataset for the LLM instruction tuning. Given the high cost of obtaining ground truth reasoning traces for complex graph-based tasks in real-world scenarios, we adopt a synthetic data generation strategy. Specifically, by leveraging graph queries from the open-source Graph-R1 dataset, we construct SSR pipeline instruction prompts, as outlined in Section~\ref{sec:method_ssr}, and queries a powerful teacher model (\textit{e.g.}, DeepSeek-R1~\cite{deepseekai2025deepseekr1incentivizingreasoningcapability}) to generate detailed chain-of-thought reasoning traces and final answers. To ensure the rigor of the training data, we introduce multi-dimensional quality control filters to identify and filter out poor-quality reasoning traces:

\begin{itemize}[leftmargin=*, noitemsep]
    \item \textbf{Subgraph Authenticity Verification}: We rigorously verify that all nodes and edges within each sampled candidate subgraph are authentic subsets of the original graph. Any instance containing hallucinated structural elements is discarded.    
    \item \textbf{Structural Diversity Evaluation}: To ensure the model explores a broad structural and textual space, we quantify the diversity of the sampled subgraphs in the candidate group $\mathbf{S}$. Specifically, we adopt an energy-based metric~\cite{velikonivtsev2024challenges} to measure the diversity of $\mathbf{S}$:
    \begin{equation}
        \text{Energy}(\mathbf{S}) = -\frac{1}{\left | \mathbf{S} \right | (\left | \mathbf{S} \right | -1)} \sum_{i \neq j} \frac{1}{D(g_i, g_j)},
    \end{equation}
    where $D(\cdot, \cdot)$ denotes the distance function between a pair of graphs. To effectively capture both topological structure and textual semantics across diverse datasets and domains, we leverage LLMs to assess the pairwise distance (with the specific prompt provided in Online Appendix~\ref{app:promot_diver}). A higher $\text{Energy}(\mathbf{S})$ value indicates greater diversity, and thus only instances exceeding a predefined energy threshold are retained.
    \item \textbf{Selection Consistency Check}: We ensure that the subgraph ultimately selected by the model for reasoning strictly belongs to the previously sampled candidate group. Any discrepancy between the selected subgraph and the available candidates leads to the exclusion of the instance.
    \item \textbf{Answer Check}: Finally, we validate the model's predicted label against the ground-truth label, retaining only instances with correct predictions through graph reasoning.
\end{itemize}
Through this rigorous distillation process, we curate a final SFT dataset comprising over 8,000 high-quality SSR-style graph reasoning demonstrations. We subsequently fine-tune the model to reconstruct the complete reasoning trajectory. Specifically, we employ a standard cross-entropy loss to minimize the discrepancy between the model's output and the high-quality synthetic traces. The objective function is defined as:
\begin{equation}
    \setlength{\abovedisplayskip}{3pt} 
    \setlength{\belowdisplayskip}{3pt} 
    \mathcal{L}_{SSR-SFT} = -\sum_{t=1}^T \log P(s_t \mid s_{<t}, \mathcal{T}(\mathcal{G}_{sub}, \mathbf{X}_{\text{sub}}, \mathcal{Y}, \mathcal{D})),
\end{equation}
where $s$ represents the entire token sequence encompassing the subgraph sampling, selection, and final graph reasoning process. This training regime ensures the model internalizes the full SSR pipeline, simultaneously enhancing its subgraph denoising and reasoning capabilities, thereby establishing a powerful foundation for subsequent reinforcement learning.

\subsection{SSR-RL}\label{sec:method_rl}
While SSR-SFT equips the model with essential graph reasoning capabilities through the SSL pipeline, specifically by learning from high-quality teacher-generated reasoning traces, these general-purpose signals might not fully enable adaptive subgraph denoising and reasoning in complex, diverse zero-shot graph reasoning settings. To further enhance these capabilities, we propose a two-stage RL framework utilizing the GRPO algorithm. GRPO is particularly well-suited to our SSR pipeline due to its group-based reward mechanism, which aligns with our strategy of sampling diverse subgraph candidates to promote both exploration and generalization. 


\subsubsection{\textbf{Stage I: Authenticity-Reinforced RLVR}}

In the first stage, we employ Authenticity-Reinforced Reinforcement Learning from Verifiable Rewards (RLVR) to reduce hallucinations during subgraph denoising, \textit{i.e.}, sampling and selecting, and strengthen the model's foundational graph reasoning ability. Unlike traditional outcome-based RLVR, which focuses solely on final answer correctness, the reward function $R_1$ is designed with nested logic to enforce the integrity and correctness of the ``Sample-Select-Reason'' process. Specifically, the reward is calculated based on three verifiable conditions, as described in Section~\ref{sec:method_sft}: the authenticity of the sampled subgraphs ($\text{Status}_{real}$), the consistency of the selected subgraph within the sampled subgraph group ($\text{Status}_{consist}$), and the correctness of the final answer ($\text{Status}_{ans}$).
\begin{equation}
    R_1 = \begin{cases} 1.0 & \text{if } \text{Status}_{real} \land \text{Status}_{consist} \land \text{Status}_{ans} \\ 0.1 & \text{if } \text{Status}_{real} \land \text{Status}_{consist} \land \neg \text{Status}_{ans} \\ 0.05 & \text{if } \text{Status}_{real} \land \neg \text{Status}_{consist} \\ 0.0 & \text{if } \neg \text{Status}_{real} \end{cases},
\end{equation}
This hierarchical mechanism is designed to incentivize the model to solve complex zero-shot graph reasoning tasks by strictly adhering to the SSR pipeline. While $\text{Status}_{ans}$ remains the ultimate objective, the intermediate rewards calculated by $\text{Status}_{real}$ and $\text{Status}_{consist}$ ensure that the model achieves the correct answer through a verifiable and consistent logical process rather than relying on shortcut heuristics.

\subsubsection{\textbf{Stage II: Denoising-Reinforced RLVR}}

Experimental observations reveal that models trained solely with Authenticity-Reinforced RLVR, while adhering to the SSR pipeline and performing authentic subgraph sampling and consistent selection for subgraph denoising, still exhibit a bias toward selecting larger subgraphs. These models tend to maximize information density at the expense of introducing excessive noise, primarily due to the lack of direct guidance for denoising. To address this, we introduce Denoising-Reinforced RLVR, which incorporates structural parsimony into the reward function. This modification explicitly penalizes structural noise, thereby guiding the model toward more effective subgraph denoising as follows:
\begin{equation}
    R_2 = \begin{cases} 1 + \lambda \cdot r_s & \text{if } R_1 = 1.0 \\ R_1 & \text{otherwise} \end{cases},
\end{equation}
where $r_s$ encourages the model to select a ``purer'' subgraph from the candidates. It is computed based on the sizes of subgraphs within the group $\mathbf{S}$:
\begin{equation}
    r_s = \frac{\rho_i}{\left | \mathbf{S} \right | - 1},
\end{equation}
where $\rho_i$ denotes the descending rank index of the selected subgraph's size within $\mathbf{S}$. The hyperparameter $\lambda$ governs the denoising intensity; while an overly small $\lambda$ may fail to induce the desired denoising effect, an excessive value risks filtering out critical information. Therefore, $\lambda$ must be carefully calibrated to ensure optimal performance. The complete GRPO-based training process for the two-stage SSR-RL framework is detailed in Online Appendix~\ref{app:implement}.

\section{Experiments}
In this section, we conduct extensive experiments to evaluate the performance of the GraphSSR framework from five perspectives: \textbf{(1)} a comparison with state-of-the-art (SOTA) baselines (Section~\ref{sec:q1}), \textbf{(2)} performance evaluation against general large reasoning models (Section~\ref{sec:q2}), \textbf{(3)} the contribution of various modeling modules (Section~\ref{exp:ablation}), \textbf{(4)} the impact of the denoising intensity hyperparameter $\lambda$ (Section~\ref{sec:q4}), and \textbf{(5)} the effectiveness of the denoising mechanism in pruning structural noise (Section~\ref{exp:case_study}).


\begin{table*}[htp!]
    \centering
    \caption{Main Results Comparison. We report Accuracy (ACC) across various tasks. The best results are indicated in \textbf{bold}, and the second-best are \underline{underlined}. Results of baselines are from the Graph-R1 paper, and ``-'' denotes results are not available.}
    \label{tab:main_results}
    \setlength{\tabcolsep}{10pt}
    \begin{tabular}{l|cc|cc|ccc|c}
        \toprule
        \multirow{2}{*}{\textbf{Method}} & \multicolumn{2}{c|}{\textbf{Cora}} & \multicolumn{2}{c|}{\textbf{WikiCS}} & \multicolumn{3}{c|}{\textbf{Products}} & \textbf{FB15K237} \\ \cmidrule{2-9} 
         & \textbf{7} & \textbf{2} & \textbf{10} & \textbf{5} & \textbf{47} & \textbf{10} & \textbf{5} & \textbf{10} \\ \midrule
        OFA & 28.65 & 56.92 & 21.20 & 35.15 & 19.37 & 30.43 & 39.31 & - \\
        GraphGPT & 44.65 & - & - & - & 18.84 & - & - & - \\
        UniGraph & 69.53 & \underline{89.74} & 43.45 & 60.23 & 38.45 & 66.07 & 75.73 & - \\
        ZeroG & 64.21 & 87.83 & 31.26 & 48.25 & 31.24 & 51.24 & 71.29 & - \\
        LLaGA & 51.85 & 62.73 & - & - & 23.10 & 34.15 & 39.72 & - \\
        GOFA-T & 70.81 & 85.73 & 71.17 & 80.93 & 54.60 & 79.33 & 87.13 & 73.59 \\
        GOFA-F & 69.41 & 87.52 & 68.84 & 80.52 & 56.13 & 80.03 & 88.34 & \textbf{80.69} \\
        Graph-R1 & \underline{71.53} & 89.08 & \underline{78.68} & \underline{86.89} & \underline{66.59} & \underline{85.72} & \underline{91.78} & 75.17 \\ \midrule
        \textbf{GraphSSR} & \textbf{72.41} & \textbf{89.81} & \textbf{79.40} & \textbf{87.93} & \textbf{68.49} & \textbf{86.46} & \textbf{91.98} & \underline{78.94} \\ \bottomrule
    \end{tabular}
\end{table*}

\subsection{Experimental Settings}

\subsubsection{\textbf{Datasets Description}}
Following the experimental setup of GOFA~\cite{kong2024gofa}, we conduct extensive experiments across a variety of graph domains, including social networks, web links, knowledge graphs, citation networks, and e-commerce. Specifically, we evaluate the zero-shot transferability of our model on node and link classification tasks using four benchmark datasets: Cora~\cite{wen2023augmenting}, WikiCS~\cite{mernyei2020wiki}, Products~\cite{he2023harnessing}, and FB15K237~\cite{liu2023one}. These datasets are strictly excluded during both the SFT and RL stages. Detailed statistics for these datasets are provided in Table~\ref{tab:datasets} of Online Appendix~\ref{app:datasets}.


\subsubsection{\textbf{Baselines and Evaluation Metrics}}
We compare GraphSSR against a diverse set of baselines that represent the current state-of-the-art in both graph-specific models and general-purpose LLMs, providing a rigorous comparison for our proposed method.

\begin{itemize}[leftmargin=*, noitemsep]
    \item \textbf{Graph-Specific Reasoning Models}: These methods are tailored specifically for zero-shot graph-based tasks, capitalizing on feature enhancement, cross-modality alignment, or graph reasoning: OFA~\cite{liu2023one}, GraphGPT~\cite{tang2024graphgpt}, UniGraph~\cite{he2024unigraph}, ZeroG~\cite{li2024zerog}, LLaGA~\cite{chen2024llaga}, GOFA~\cite{kong2024gofa}, and Graph-R1~\cite{wu2025graph}.
    \item \textbf{General Large Reasoning Models}: These models are optimized for broad cognitive tasks. To handle graph-based tasks, they require textual serialization to perform inference: DeepSeek-R1-distilled-Qwen2.5-14B~\cite{deepseekai2025deepseekr1incentivizingreasoningcapability}, Qwen3-14B (Thinking)~\cite{qwen3technicalreport}, Ministral-3-14B-Reasoning-2512~\cite{liu2026ministral}, and DeepSeek-R1-0528~\cite{deepseekai2025deepseekr1incentivizingreasoningcapability}.
\end{itemize}
We use \textbf{Accuracy} as the primary metric to evaluate performance on both node and link classification tasks in a zero-shot setting.

\subsubsection{\textbf{Implementation Details}}

We employ DeepSeek-R1-distilled-Qwen2.5-14B as the foundational backbone for all experiments. During the SSR-SFT phase, we utilize the LlamaFactory framework~\cite{zheng2024llamafactory} to fine-tune the model for 3 epochs with a learning rate of $1.0 \times 10^{-5}$ and a batch size of 64. For the subsequent SSR-RL phase, we leverage the verl~\cite{sheng2024hybridflow} framework, training the model for 4 epochs with a more conservative learning rate of $1.0 \times 10^{-6}$ and a batch size of 64. In the Denoising-Reinforced RLVR, the denoising intensity hyperparameter $\lambda$ is set to 0.1. All computational tasks, including data synthesis and model training, are conducted on a cluster of 8 NVIDIA H100 GPUs.

\subsection{Main Results}\label{sec:q1}

\begin{table*}[htp!]
    \centering
    \caption{Comparison between \textbf{GraphSSR} and Large Reasoning Models (LRMs). Best in \textbf{bold}; second best \underline{underlined}.}
    \label{tab:performance_llm}
    \begin{tabular}{lcccc}
        \toprule
        \textbf{Methods} & \textbf{Cora (7)} & \textbf{WikiCS (10)} & \textbf{Products (47)} & \textbf{FB15K237 (10)} \\
        \midrule
        DeepSeek-R1-0528  & \underline{69.59} & 76.96 & \textbf{70.17} & \textbf{84.12} \\
        \midrule
        Qwen3-14B (Thinking)  & 65.15 & \underline{78.56} & 66.54 & 77.86 \\
        Ministral-3-14B-Reasoning-2512  & 69.30 & 75.38 & 59.89 & 77.67 \\
        DeepSeek-R1-distilled-Qwen2.5-14B & 68.80 & 75.26 & 66.09 & 73.49 \\
        \midrule
        \textbf{GraphSSR}     & \textbf{72.41} & \textbf{79.40} & \underline{68.49} & \underline{78.94} \\
        \bottomrule
    \end{tabular}
\end{table*}

Table~\ref{tab:main_results} summarizes the performance comparison between our proposed GraphSSR and several state-of-the-art (SOTA) baselines across various benchmark datasets. The values beneath each dataset header denote the number of candidate categories, where a larger number indicates increased task complexity.

As illustrated in Table~\ref{tab:main_results}, GraphSSR achieves SOTA performance on the majority of tasks. For instance, on the 7-category Cora and 47-category Products node classification tasks, our method achieves 72.41\% and 68.49\% accuracy, respectively, marking a significant improvement over baselines. While GOFA-F remains the leader on the FB15K237 link classification task, our method effectively narrows the performance gap, demonstrating strong generalizability across different graph-based tasks.

A particularly noteworthy observation is GraphSSR's performance on the Products dataset. Specifically, as the number of candidate categories increases from 5 to 47, the performance gap between our method and the runner-up, Graph-R1, widens significantly. At the maximum 47-category setting—the most challenging configuration—our method outperforms Graph-R1 by nearly 2\% (68.49\% vs. 66.59\%). This trend can be attributed to the inherent nature of the Products dataset, where labels exhibit substantial semantic overlap (\textit{e.g.}, ``Kitchen \& Dining'' and ``Grocery \& Gourmet Food''). As the label space expands, the increased reasoning complexity demands that the model extract highly precise and relevant structural information to discern subtle semantic nuances. While previous methods often succumb to distracting noise in such dense environments, our method effectively prunes task-irrelevant nodes and edges. Such superior and adaptive denoising capability allows our method to maintain high reasoning accuracy even as the task difficulty scales, underscoring the necessity of subgraph denoising for zero-shot graph reasoning in complex scenarios.

\subsection{Comparison with Large Reasoning Models} \label{sec:q2}

To comprehensively evaluate the effectiveness of \textbf{GraphSSR}, we compare it against several state-of-the-art Large Reasoning Models (LRMs). Specifically, we selected three open-source reasoning models of equivalent parameter scales, alongside the large-scale DeepSeek-R1 as a high-end benchmark. The results, summarized in Table~\ref{tab:performance_llm}, demonstrate the clear superiority of our method in integrating structural graph information with specialized graph reasoning.

As shown in the table, our method consistently outperforms all same-scale baselines. This performance gap indicates that the general-purpose knowledge and native reasoning capabilities of LLMs, while powerful, are insufficient for addressing complex zero-shot graph reasoning tasks. Without an explicit mechanism to process graph topology and mitigate structural noise, these models struggle to identify the most relevant information within a complex graph context.

Notably, our method even outperforms the full-parameter DeepSeek-R1-0528 on the Cora and WikiCS datasets. Despite DeepSeek-R1-0528 possessing a substantially larger parameter scale and superior general reasoning capabilities, its performance on these specific graph tasks is hindered by the absence of an effective subgraph denoising mechanism. These observations provide strong evidence that our proposed Sample-Select-Reason pipeline and two-stage RL strategy are essential for bridging the gap between general linguistic reasoning and graph reasoning.

\subsection{Ablation Study}\label{exp:ablation}

To evaluate the contribution of each core component in GraphSSR, we conduct ablation studies across various datasets. We consider four model variants: (1) \textit{\textbf{w/o SSR pipeline}}, which omits our Sample-Select-Reason pipeline and directly performs reasoning on the raw, task-agnostic subgraph after standard SFT and RLVR training; (2) \textit{\textbf{w/o RL}}, which utilizes the SSR pipeline during SSR-SFT but excludes the entire SSR-RL process; (3) \textit{\textbf{w/o Authenticity-Reinforced RLVR}}, which follows the full pipeline but specifically excludes the Authenticity-Reinforced RLVR; and (4) \textit{\textbf{w/o Denoising-Reinforced RLVR}}, which similarly excludes the Denoising-Reinforced RLVR.

\begin{figure}[t]
    \centering
    \includegraphics[width=0.85\linewidth]{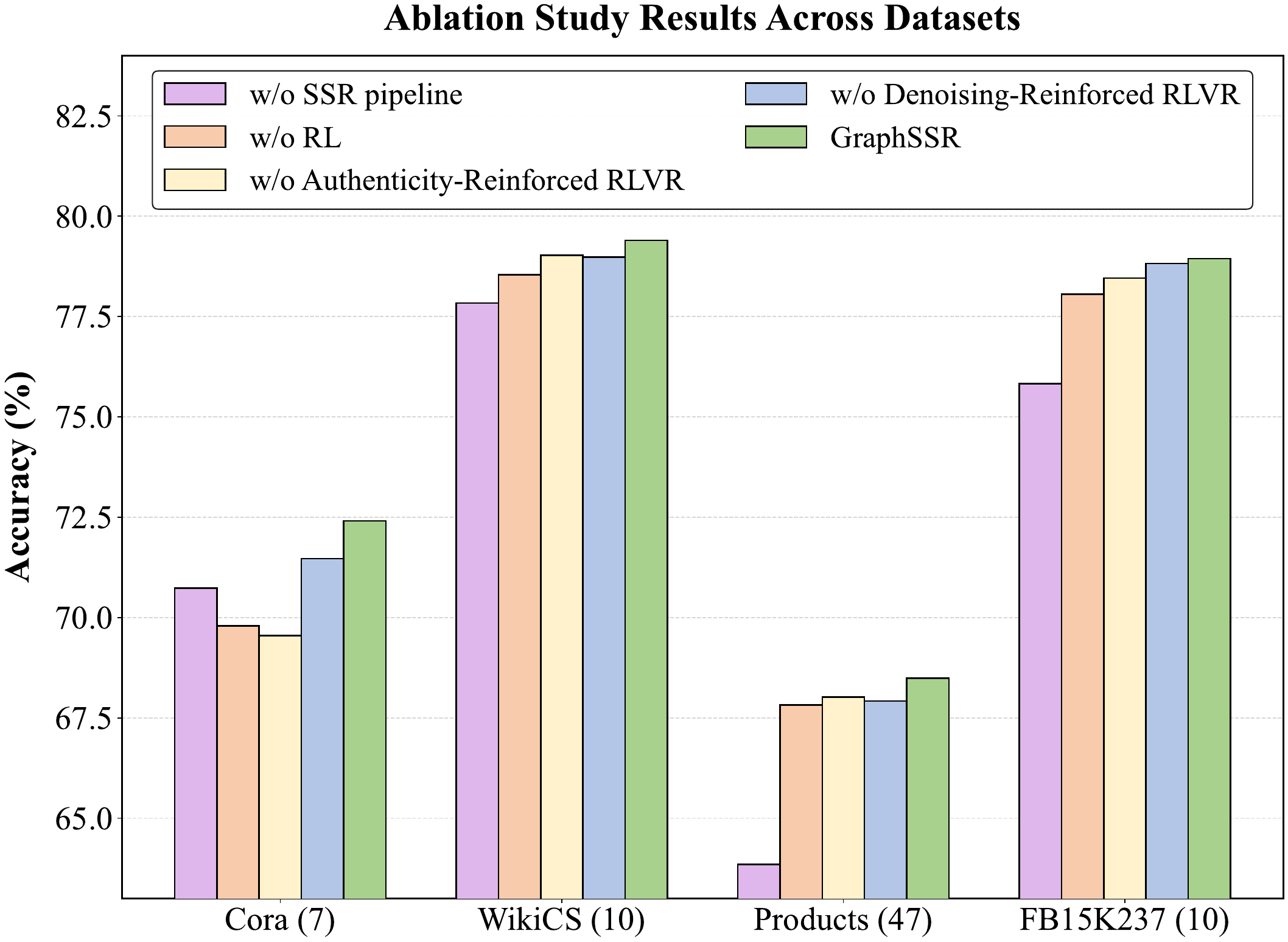} 
    \Description{Ablation study of \textbf{GraphSSR} on different datasets.}
    \caption{Ablation study of \textbf{GraphSSR} on different datasets.}
    \label{fig:ablation}
\end{figure}

The experimental results, as illustrated in Figure~\ref{fig:ablation}, consistently demonstrate the effectiveness of our proposed modules. Notably, the removal of the SSR pipeline leads to the most substantial performance degradation across the majority of datasets. This decline highlights that without the adaptive sampling and selecting phases, the model is forced to process task-agnostic subgraphs that often contain task-irrelevant nodes and edges. The inability to filter this structural noise directly impedes the model's reasoning qualities, confirming that the SSR pipeline is fundamental to zero-shot graph reasoning. Furthermore, the performance drop observed in the w/o RL setting demonstrates that SSR-SFT alone is insufficient for the model to fully internalize the complex strategies required for effective subgraph denoising and reasoning. The RL phase is essential for bridging the gap between simply mimicking the teacher's reasoning traces and autonomously exploring optimal traces in the subgraph space.

Furthermore, both Authenticity-Reinforced RLVR and Denoising-Reinforced RLVR contribute critically to the final performance, albeit through different mechanisms. Specifically, the absence of Authenticity-Reinforced RLVR weakens the model’s ability to identify task-relevant, authentic structural information and limits its fundamental reasoning capabilities. Conversely, omitting Denoising-Reinforced RLVR results in a lack of denoising capability. For a more intuitive demonstration of these effects, we provide detailed case studies in Online Appendix~\ref{app:case_rl}.


\subsection{Parameter Sensitivity} \label{sec:q4}

\begin{figure}[t]
    \centering
    \includegraphics[width=\linewidth]{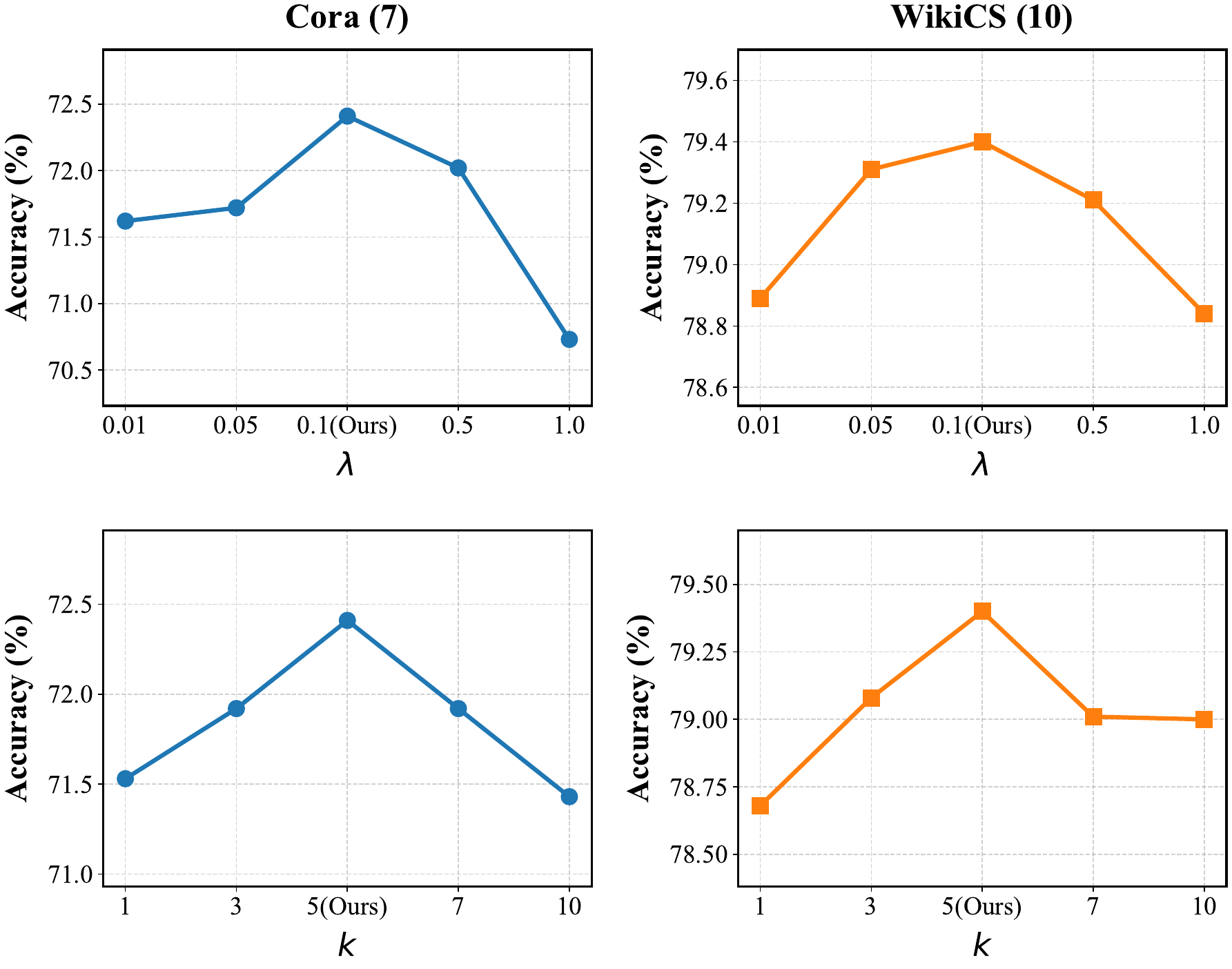} 
    \Description{Parameter Sensitivity of the reward weight $\lambda$ and candidate group size $k$.}
    \caption{Parameter Sensitivity of the reward weight $\lambda$ and candidate group size $k$.}
    \label{fig:parameter}
    \vspace{-0.5cm}
\end{figure}


In this section, we investigate the sensitivity of GraphSSR to two core hyperparameters: the reward weight $\lambda$, which governs the denoising intensity during the Denoising-Reinforced RLVR stage, and the candidate group size $k$, which dictates the scale of the sampled candidate subgraph pool.


As illustrated in the top row of Figure~\ref{fig:parameter}, varying $\lambda$ from $0.01$ to $1.0$ yields a consistent ``inverted U-shape'' trend across both datasets. When $\lambda$ is set to a relatively low value (\textit{e.g.}, $0.01$), the performance is suboptimal. This is primarily because a small $r_s$ weight fails to impose sufficient pressure on the model to prune noisy structures, leading the model to suffer from the interference of noisy nodes and edges. As $\lambda$ increases, we observe a steady improvement in accuracy, reaching a peak at $\lambda=0.1$. This upward trend validates that introducing a properly weighted denoising reward successfully guides the model toward selecting ``purer'' and more relevant subgraphs, which enhances its reasoning quality. However, as $\lambda$ continues to increase beyond the optimal threshold, the performance starts to decline on both datasets. This phenomenon suggests that an excessively high penalty may lead to an ``over-pruning'' issue, where the model becomes overly biased toward structural brevity at the expense of losing critical semantic information required for correct classification. 

Furthermore, the bottom row of Figure~\ref{fig:parameter} depicts the impact of $k$. As $k$ increases, the performance of the model initially improves because a larger $k$ provides the necessary diversity and selection space for the SSR pipeline to effectively identify task-relevant structures. However, as $k$ exceeds 5, the performance plateaus or even declines, as an excessively large pool introduces redundant noisy subgraphs that can distract the model’s selection and reasoning. This confirms that GraphSSR's efficacy stems from identifying the optimal equilibrium between purity and informativeness rather than brute-force computational scaling. For practical deployment, we found $k=5$ to be the optimal trade-off between performance and efficiency.


\begin{table*}[t]
    \centering
    \caption{Quantitative evaluation of denoising quality and structural trade-offs.}
    \label{tab:denoising_analysis}
    \begin{tabular}{llcccc}
    \toprule
    \textbf{Metric} & \textbf{Strategy} & \textbf{Cora (7)} & \textbf{WikiCS (10)} & \textbf{Products (47)} & \textbf{FB15K237 (10)} \\
    \midrule
    \multirow{3}{*}{Noise Ratio (\%)} 
    & Raw Subgraph       & 23.86 & 68.19 & 41.25 & 56.95 \\
    & Smaller Candidates & 12.29 & 35.05 & 30.38 & 49.91 \\
    & \textbf{GraphSSR}           & 13.91 & 35.91 & 32.61 & 45.22 \\
    \midrule
    \multirow{3}{*}{Avg. Node Count} 
    & Raw Subgraph       & 18.6  & 27.3  & 18.0  & 32.1  \\
    & Smaller Candidates & 2.9   & 3.0   & 3.0   & 3.6   \\
    & \textbf{GraphSSR}           & 5.9   & 6.8   & 7.4   & 8.3   \\
    \bottomrule
    \end{tabular}
\end{table*}

\subsection{Case Study}\label{exp:case_study}

\subsubsection{\textbf{Reasoning Trajectory Analysis Under Adaptive Subgraph Denoising}}\label{exp:case_qualitative}

To better understand how GraphSSR performs subgraph denoising and enables correct graph reasoning, we conduct a qualitative analysis on a specific zero-shot node classification task from the Cora dataset. In this example, the target node 11's textual content is: ``\textit{Hierarchical logistic belief networks for linear model selection, using Gibbs sampling for parameter learning}'', with the ground-truth label ``\textbf{Neural Networks}''. Within its neighborhood, there are semantically consistent nodes that support correct reasoning (\textit{e.g.}, node 13, node 17), as well as detrimental noisy nodes (\textit{e.g.}, node 9, node 14) whose textual attributes pertain to another label, ``\textbf{Probabilistic Methods}'', which is somewhat connected to the target node but incorrect.

In the absence of our proposed SSR pipeline and the associated subgraph denoising training strategy, the model takes as input a uniform, task-agnostic, and unfiltered ego-subgraph (\textit{i.e.}, the $h$-hop subgraph commonly used in existing work). Without a denoising mechanism, the descriptions of noisy neighbors, such as ``EM algorithm'', can overwhelm the specific attributes of the target node. Despite the presence of semantically consistent nodes, the interference from noisy information misleads the model into producing an incorrect prediction: ``Probabilistic Methods''. In contrast, GraphSSR demonstrates a decisive strategic advantage. By effectively selecting a task-adaptive subgraph $Subgraph\_2$—which retains relevant neighbors (node 13 and node 17) while pruning noisy ones (node 9 and node 14)—our method strikes an optimal balance between purity and informativeness. This task-adaptive, denoised context empowers the reasoning module to focus on the precise semantics of the target node, \textit{i.e.}, ``networks'', ultimately yielding the correct classification: ``Neural Networks''. The complete case study, model output, and additional analysis of other model variants are provided in Online Appendix~\ref{app:case_ssr}.



\subsubsection{\textbf{Graph Analysis Under Adaptive Subgraph Denoising}}

To quantitatively validate the effectiveness and rationality of GraphSSR's denoising mechanism, we evaluate the candidate subgraphs across 100 randomly sampled cases for each dataset using two key structural metrics: (1) \textbf{Noise Ratio (\%)}, which measures the proportion of noisy neighbor nodes within the subgraph. Due to the inherent absence of ground-truth noise labels in benchmarks, we design a task-specific heuristic metric where nodes exhibiting low LLM-judged semantic similarity to the target node are identified as noise. (2) \textbf{Avg. Node Count}, which quantifies the average structural scale (total number of nodes, including the target and its neighbors) of the subgraphs to reflect structural parsimony. Under this framework, we systematically evaluate GraphSSR against two baseline configurations: (i) \textbf{Raw Subgraph}, which represents the initial, unfiltered subgraph commonly used in existing strategies, and (ii) \textbf{Smaller Candidates}, which isolates all candidate subgraphs sampled during the \textit{Sample Phase} that possess a total node count strictly smaller than the final selection of GraphSSR.

As summarized in Table~\ref{tab:denoising_analysis}, the initial Raw Subgraph suffers from severe noise across all datasets, peaking at 68.19\% on WikiCS. GraphSSR successfully compresses the average node count while dramatically suppressing the noise ratio. Crucially, this simultaneous reduction in subgraph size and improvement in task performance provides strong empirical evidence that more information ``does not equate to better reasoning'' in graph reasoning tasks. This trend further underscores GraphSSR's ability to effectively prune large-scale structural noise through its SSR pipeline and denoising strategy.

Conversely, while the Smaller Candidates achieve slightly lower noise ratios on certain datasets due to their highly restricted size, they suffer from a severe \textit{over-pruning} issue. Their average node count drops precipitously to approximately 3 nodes (consisting of only the target node and around two neighbors), which starves the model of necessary context and consequently sabotages reasoning accuracy. Notably, statistical analysis reveals that GraphSSR \textbf{does not select} the smallest candidate in \textbf{89.75\%} of cases. This empirical evidence demonstrates that our verifiable reward does not naively prefer smaller subgraphs, but instead successfully achieves an optimal trade-off between structural purity and informativeness, ensuring essential signals are preserved while noise is removed.

\section{Related Work}

\subsection{Integrating LLMs with Graph Modules}

This line of research explores the synergy between LLMs and specialized graph-specific modules, such as GNNs, to integrate the semantic depth of LLMs with the structural inductive biases of GNNs. It can primarily be divided into two aspects:

\paragraph{\textbf{LLM as Enhancer.}}

This category of methods leverages the superior semantic reasoning capabilities and inherent knowledge of LLMs to enrich graph data, providing high-quality features for downstream GNNs. Representative methods such as OFA~\cite{liu2023one}, UniGraph~\cite{he2024unigraph}, and ZeroG~\cite{li2024zerog} utilize LLMs to project diverse node attributes into a unified semantic space, facilitating cross-domain feature alignment. To address scenarios with missing or sparse textual information, TANS~\cite{wang2025can} employs LLMs to synthesize high-quality textual descriptions from raw graph structures. Beyond static enhancement, other studies introduce dynamic refinement and knowledge transfer: DAS~\cite{thapaliya2025semantic} implements a feedback-driven loop where an LLM iteratively refines node text descriptions guided by GNN feedback to better align semantics with graph topology, and PKD~\cite{wei2025preference} adopts a preference-driven knowledge distillation strategy to transfer task-specific expertise from LLMs to GNNs for enhanced few-shot classification. Despite their effectiveness, a primary limitation of these methods is their persistent reliance on GNN training, which often leads to suboptimal transferability and performance in zero-shot settings.

\paragraph{\textbf{LLM as Predictor.}}

In this paradigm, LLMs serve as the final reasoning engine, where graph-specific structural information is first encoded by a graph module and then integrated into the LLM's input space for prediction. A core group of methods focuses on aligning graph topological features with the LLM's embedding space; for instance, LLaGA~\cite{chen2024llaga} transforms neighborhood structures into sequential representations via predefined templates, whereas GraphGPT~\cite{tang2024graphgpt}, GraphTranslator~\cite{zhang2024graphtranslator}, and TEA-GLM~\cite{wang2024llms} utilize multi-stage alignment or translation modules to bridge the modality gap. To endow LLMs with intrinsic GNN capabilities, GOFA~\cite{kong2024gofa} integrates GNN layers into the LLM architecture, while UniGTE~\cite{wang2025unigte} recovers node-permutation invariance like GNNs by modifying the attention mechanism within the LLM. Despite their progress, this ``encoder-then-reasoner'' paradigm often suffers from modality decoupling, where the reliance on complex, domain-specific alignment pretraining limits their adaptability to unseen label spaces and hinders universal zero-shot generalization.

\subsection{Vanilla LLM-based Graph Reasoning}

Another direction explores the inherent reasoning capabilities of LLMs to solve graph-based tasks directly through natural language, obviating the need for external graph-specific modules. Earlier methods convert graph topologies into natural language prompts, enabling pretrained LLMs to perform zero-shot or few-shot reasoning without parameter updates~\cite{hu2023beyond, wang2023graph, li2025large, sun2025graphicl}. While they capitalize on the general-purpose reasoning of LLMs, the lack of domain-specific adaptation often leads to suboptimal performance on tasks requiring deep structural understanding. To overcome these limitations, subsequent research has shifted from frozen prompting to fine-tuning. Specifically, InstructGraph~\cite{wang2024instructgraph} utilizes structured reasoning tasks and preference alignment to guide graph-specific outputs, while InstructGLM~\cite{ye2024language} employs multi-prompt training across diverse tasks to improve cross-task generalization. Unlike traditional instruction tuning, recent studies have attempted to propose unique training paradigms. LangGFM~\cite{lin2024langgfm} introduces self-supervised alignment via topology and feature-masked autoencoding, whereas GRIP~\cite{feng2025grip} leverages LoRA fine-tuning to internalize graph structures within model parameters, enabling efficient reasoning without external structural inputs during inference. Furthermore, Graph-R1~\cite{wu2025graph}, which is most closely related to our method, employs reinforcement learning to incentivize zero-shot graph learning by driving explicit, long chain-of-thought reasoning through task-specific templates. However, Graph-R1 relies on a task-agnostic, fixed $h$-hop subgraph extraction strategy, which inevitably introduces task-irrelevant structural noise that interferes with the reasoning process. To address this, our method proposes a subgraph denoising SSR pipeline coupled with a two-stage reinforcement training strategy. Distinct from existing methods, our method performs reasoning over purified, task-relevant subgraphs, thereby narrowing the receptive field to essential components and significantly enhancing overall performance.

\section{Conclusion}

In this paper, we address the critical challenge of structural noise in zero-shot graph reasoning by proposing GraphSSR, a novel framework that shifts the modeling paradigm from traditional ``one-size-fits-all'' task-agnostic subgraph extraction to an adaptive Sample-Select-Reason (SSR) pipeline with an explicit subgraph denoising mechanism. In this way, GraphSSR enables Large Language Models to autonomously explore a diverse space of candidate subgraphs, filter out task-irrelevant nodes and edges, and execute high-fidelity reasoning on a denoised structural context. To enable the model to execute the SSR pipeline, we integrate a rigorous SSR-SFT data synthesis strategy with a two-stage reinforcement learning framework, SSR-RL, which utilizes Authenticity-Reinforced RLVR to eliminate hallucinations during subgraph sampling and selecting while solidifying reasoning capabilities, and Denoising-Reinforced RLVR to penalize structural noise through a subgraph size-based reward mechanism. By incentivizing the model to achieve correct predictions using the concise and relevant structural information, GraphSSR achieves state-of-the-art performance across multiple zero-shot benchmarks, providing a robust and effective solution for generalizing graph reasoning to unseen domains and label spaces.



\begin{acks}
The work of Liang Zhang is supported/funded by the Guangzhou-HKUST(GZ) Joint Funding Program (No. 2024A03J0630). The work of Yuan Zuo is partially supported by the National Natural Science Foundation of China (72571019, 72542016).
\end{acks}

\bibliographystyle{unsrt}
\bibliography{sample-base}

\appendix

\section{Stability and Robustness Analysis}\label{app:robustness_analysis}

To evaluate the stability and robustness of GraphSSR, we conduct five independent runs using distinct random seeds. The experimental results, reported as mean $\pm$ standard deviation in Table~\ref{tab:performance_stability}, demonstrate that GraphSSR consistently maintains low variances (ranging from $\pm 0.08\%$ to $\pm 0.39\%$). This remarkable stability firmly verifies that GraphSSR's performance gains over state-of-the-art baselines (e.g., Graph-R1) are statistically robust and not a byproduct of stochastic experimental noise, even when deployed in challenging, high-cardinality classification tasks such as Products (47).

\begin{table*}[t]
    \centering
    \caption{Performance stability across multiple runs (reported as mean $\pm$ standard deviation).}
    \label{tab:performance_stability}
    \begin{tabular}{cccccccc}
    \toprule
    \textbf{Cora (7)} & \textbf{Cora (2)} & \textbf{WikiCS (10)} & \textbf{WikiCS (5)} & \textbf{Products (47)} & \textbf{Products (10)} & \textbf{Products (5)} & \textbf{FB15K237 (10)} \\
    \midrule
    72.32 $\pm$ 0.39 & 89.87 $\pm$ 0.20 & 79.34 $\pm$ 0.14 & 87.89 $\pm$ 0.16 & 68.41 $\pm$ 0.18 & 86.48 $\pm$ 0.11 & 91.90 $\pm$ 0.09 & 78.75 $\pm$ 0.08 \\
    \bottomrule
    \end{tabular}
\end{table*}

\section{Performance Evaluation on the Link Prediction Task}\label{app:link_prediction}

To further evaluate the capability of GraphSSR on different tasks, we conduct additional experiments on the link prediction task using the unseen Cora and Products datasets in training. As shown in Table~\ref{tab:unseen_link_prediction}, GraphSSR consistently and significantly outperforms the state-of-the-art Graph-R1 baseline. This improvement demonstrates that the adaptive SSR pipeline and denoising mechanism are highly effective for structural reasoning in the link prediction task, reinforcing the robustness of GraphSSR.

\begin{table}[h]
    \centering
    \caption{The comparison of accuracy on the link prediction task across unseen datasets.}
    \label{tab:unseen_link_prediction}
    \begin{tabular}{lcc}
        \toprule
        \textbf{Method} & \textbf{Cora} & \textbf{Products} \\
        \midrule
        Graph-R1 & 91.63 & 81.14 \\
        \textbf{GraphSSR} & \textbf{93.50} & \textbf{91.64} \\
        \bottomrule
    \end{tabular}
\end{table}

\section{In-Depth Analysis of the SSR-SFT Stage}\label{app:ssr_sft_analysis}

In this section, we present comprehensive evaluations to systematically validate our design choices within the Subgraph-Sample-Reason Supervised Fine-Tuning (SSR-SFT) stage, encompassing both diversity filter variants and training data dimension analyses.

\subsection{Ablation and Variants of the Structural Diversity Evaluation Filter}\label{app:filter_ablation_sec}

To verify the necessity and validity of our proposed data-filtering pipeline, we examine the behavior of the Structural Diversity Evaluation filter under two distinct settings: (i) completely removing the filter (\textbf{SFT\_w/o\_Structural\_Diversity\_Evaluation}), and (ii) replacing it with three deterministic graph-similarity heuristics. The deterministic baselines are defined as follows:
\begin{itemize}
    \item \textbf{SFT\_with\_WL\_kernel\_filter}: Utilizes Weisfeiler-Lehman (WL) Graph Kernels~\cite{shervashidze2011weisfeiler} to measure purely structural topology similarity.
    \item \textbf{SFT\_with\_text\_sim\_filter}: Employs BGE-M3~\cite{chen2024bge} to vectorize node text and computes the average cosine similarity between subgraphs.
    \item \textbf{SFT\_with\_hybrid\_filter}: Takes the arithmetic average of both the WL kernel and BGE-M3 similarity metrics.
\end{itemize}
We replace our proposed Structural Diversity Evaluation filter with these deterministic alternatives or remove it to regenerate SFT data of the identical volume for a fair comparison. To cleanly isolate the direct impact of the data filtering process, we fine-tune the base model within the SSR-SFT stage—bypassing the subsequent SSR-RL stage—and evaluate the performance of these SFT-only variants. The empirical results are summarized in Table~\ref{tab:filter_ablation}.

\begin{table*}[t]
    \centering
    \caption{Results on the ablation and variants of the Structural Diversity Evaluation filter.}
    \label{tab:filter_ablation}
    \begin{tabular}{lcccc}
        \toprule
        \textbf{Filtering Configuration} & \textbf{Cora (7)} & \textbf{WikiCS (10)} & \textbf{Products (47)} & \textbf{FB15K237 (10)} \\
        \midrule
        SFT\_w/o\_Structural\_Diversity\_Evaluation & 69.30 & 77.94 & 65.96 & 77.25 \\
        SFT\_with\_WL\_kernel\_filter                     & 68.31 & 78.22 & 66.66 & 77.26 \\
        SFT\_with\_text\_sim\_filter                      & 68.07 & 78.13 & 67.09 & 77.22 \\
        SFT\_with\_hybrid\_filter                         & 68.81 & 78.29 & 67.06 & 77.28 \\
        \textbf{SFT\_GraphSSR}                      & \textbf{69.80} & \textbf{78.54} & \textbf{67.83} & \textbf{78.07} \\
        \bottomrule
    \end{tabular}
\end{table*}

As shown in the results, completely removing the Structural Diversity Evaluation filter leads to a consistent performance decline across all four benchmarks. This provides strong empirical evidence that a diversity-driven filter is a prerequisite for the model to effectively execute subsequent adaptive denoising. 

Furthermore, GraphSSR with the Structural Diversity Evaluation filter leveraging the LLM-based diversity score (\textbf{SFT\_GraphSSR}) consistently outperforms all deterministic variants. While WL kernels and BGE-M3 text similarity provide useful isolated structural and semantic signals respectively, they struggle to capture the complex, task-specific cross-modal interactions between graph topology and textual nuances in a zero-shot setting. In contrast, the LLM-based metric effectively integrates these two dimensions, thereby confirming the superiority and validity of our filtering strategy.

\subsection{Impact of Data Curation Dimensions: Quality vs. Quantity}\label{app:teacher_data_impact}

We further investigate the empirical impact of the SSR-SFT stage across two vital data curation dimensions: the quality of the teacher model generating the reasoning trajectories, and the quantity of the synthesized training data. Specifically, we replace our default powerful teacher \texttt{DeepSeek-R1-0528} with a weaker counterpart, \texttt{DeepSeek-R1-distilled-Qwen2.5-32B} (\textbf{SFT\_Teacher\_32B}), and separately restrict the training process to use only 20\% of the synthetic data scale (\textbf{SFT\_20\%\_Data}). The comparative results are reported in Table~\ref{tab:teacher_data_ablation}.

\begin{table*}[t]
    \centering
    \caption{Impact analysis regarding the training data configurations.}
    \label{tab:teacher_data_ablation}
    \begin{tabular}{lcccc}
        \toprule
        \textbf{Configuration} & \textbf{Cora (7)} & \textbf{WikiCS (10)} & \textbf{Products (47)} & \textbf{FB15K237 (10)} \\
        \midrule
        SFT\_Teacher\_32B (Quality)   & \textbf{69.88} & 75.50 & 62.21 & 64.76 \\
        SFT\_20\%\_Data (Quantity)        & 68.61          & 78.08 & 67.50 & 76.66 \\
        \textbf{SFT\_GraphSSR} & 69.80          & \textbf{78.54} & \textbf{67.83} & \textbf{78.07} \\
        \bottomrule
    \end{tabular}
\end{table*}

Regarding teacher quality, substituting the original teacher with the distilled 32B variant leads to significant performance drops on complex datasets (e.g., -5.62\% on Products and -13.31\% on FB15K237), confirming that a powerful teacher is essential for generating high-fidelity reasoning traces in intricate graph environments. Regarding data scale, training with only 20\% of the synthetic data also results in a noticeable decline across all benchmarks; however, it consistently outperforms the model trained with a weaker teacher on complex benchmarks (e.g., 76.66\% vs. 64.76\% on FB15K237). 

This indicates that for zero-shot graph reasoning, the quality of reasoning traces is a more dominant factor than data volume. Therefore, under computational constraints, prioritizing a stronger teacher model to generate high-quality samples—even at a smaller scale—is a more effective strategy for enhancing the generalization of GraphSSR.

\section{Alternative Design Analysis of the Selection Mechanism}\label{app:ssr_pipeline_alternative}

To evaluate the efficacy of our adaptive selection strategy within the SSR pipeline, we conduct a comparative analysis against a static, similarity-based heuristic baseline designated as \textbf{Graph-R1-Select}. This alternative design tests whether a hard-filtered, rule-based pruning strategy can achieve comparable denoising effects without adaptive reinforcement. Specifically, in \textbf{Graph-R1-Select}, neighbor nodes that exhibit low semantic similarity to the target node—as evaluated directly via the LLM's textual comprehension—are statically pruned from the complete subgraph prior to reasoning. We contrast this alternative strategy against both the vanilla \textbf{Graph-R1} (which utilizes unfiltered complete subgraphs) and our full \textbf{GraphSSR} framework.

\begin{table*}[t]
    \centering
    \caption{Performance comparison between GraphSSR and the static selecting baseline.}
    \label{tab:ssr_pipeline_comparison}
    \begin{tabular}{lcccc}
        \toprule
        \textbf{Method} & \textbf{Cora (7)} & \textbf{WikiCS (10)} & \textbf{Products (47)} & \textbf{FB15K237 (10)} \\
        \midrule
        Graph-R1-Select & 70.93 & 76.99 & 62.59 & 76.08 \\
        Graph-R1                    & 71.53 & 78.68 & 66.59 & 75.17 \\
        \textbf{GraphSSR}           & \textbf{72.41} & \textbf{79.40} & \textbf{68.49} & \textbf{78.94} \\
        \bottomrule
    \end{tabular}
\end{table*}

As demonstrated in Table~\ref{tab:ssr_pipeline_comparison}, this static selecting baseline often leads to performance degradation compared to Graph-R1—most notably a 4.0\% drop on the Products dataset—indicating that simple heuristics lack the task-specific understanding required for effective graph denoising and may inadvertently remove critical information. In contrast, GraphSSR’s SSR mechanism effectively balances denoising and information preservation, consistently achieving superior performance across all benchmarks and underscoring the necessity of our proposed framework.

\section{Online Appendix}\label{app:online_appendix}

Due to space limitations, the remaining appendix is provided online and can be accessed at \url{https://github.com/mysteriouslfz/GraphSSR/blob/master/appendix/Online_Appendix.pdf}.

\clearpage

\section{Prompt Templates}

\subsection{Sample-Select-Reason Pipeline}\label{app:prompt_ssr}

\begin{table*}[hbp!]
    \begin{tcolorbox}[
        title=\textbf{\textcolor{white}{A Prompt Example on Node Classification for SSR pipeline}},
        colframe=black,
        coltitle=black,
        boxrule=1.0pt,
        arc=1mm,
        auto outer arc,
    ]
    \textcolor{red}{\#\#\#\# Task Description and Initialized Subgraph}\\

    \textbf{Task Description}:\\
    You are a graph analysis expert. Given a complete subgraph structure, which includes the description of the central node(s), its neighboring nodes, and their connection relationships, please follow the following steps to operate the analysis:\\
    
    \textbf{Complete subgraph}:\\
    - Central node ID(s): nodex\\
    - Node texts (central node(s) and neighboring nodes): [nodex: node description]\\
    - Connection relationships: [<nodex, nodey>]\\

    \textcolor{red}{\#\#\#\# Sampling Phase}\\
    
    \textbf{Step description}:
    
    1. \textbf{Subgraph Sampling}:\\
      - In order to complete the Node Classification task of the central node(s), you need to first sample \textbf{5} subgraphs from the complete graph that are most suitable for the current task from this subgraph.\\
      - These subgraphs must \textbf{keep all central node(s)} as mentioned in the complete subgraph, they can include no, partial or all one-hop or two-hop neighbors of the central node(s).\\
      - You must ensure that the sampled subgraphs \textbf{actually exist} and do not create fictitious nodes and edges.\\
      - There must be a subgraph that consists \textbf{only the central node(s)}.\\
      - There should also be a subgraph, where as many neighboring nodes as possible can \textbf{semantically support the central node(s)}. If there is no such subgraph, do not fabricate one.\\
      - You must ensure that the \textbf{information differences} between these subgraphs are as significant as possible, so that these subgraphs can represent different semantic and structural meanings. The sizes of the subgraphs should also be as different as possible.\\

    \textcolor{red}{\#\#\#\# Selection Phase}\\
    
    2. \textbf{Subgraph Choosing}:\\
      From the sampled subgraphs, choose one subgraph that is \textbf{best suited for the current task} and use it for subsequent reasoning.\\

    \textcolor{red}{\#\#\#\# Reasoning Phase}\\
    
    3. \textbf{Node Classification Reasoning}:\\
      - Repeat and verify the subgraph you have chosen, ensuring that the chosen subgraph you analyze and output \textbf{is consistent with} the sampled subgraphs.\\
      - Analyze the chosen subgraph independently. \textbf{Only consider the nodes and edges of the current subgraph}. The obtained answer is entirely based on the chosen subgraph and is independent of other sampled subgraphs and the complete subgraph.\\
      - You must \textbf{treat the current subgraph as a whole} for reasoning, taking into account the information of all nodes in this subgraph, not just the information of the central node(s).\\
      - Provide the Node Classification answer of the central node(s) obtained through reasoning, the answer must be \textbf{selected strictly} from the following Node Classification options. Each option is enclosed within < >, but when you answer, do not include < >.\\
    
    4. \textbf{Node Classification Options}: <Theory>, <Probabilistic Methods>, <Genetic Algorithms>, <Reinforcement Learning>, <Case-Based>, <Neural Networks>, <Rule Learning>.
    
    \end{tcolorbox}
\end{table*}

In this section, we provide a prompt example for the zero-shot node classification task and detail the prompt template design for the \textbf{SSR} pipeline. The pipeline begins with the \textbf{Task Description and Initialized Subgraph}, where we define the model's persona as a ``graph analysis expert'' and provide the raw subgraph data. This stage establishes a comprehensive context, ensuring the model captures both node semantics and explicit topological connections as a foundation for subsequent reasoning.

In the \textbf{Sample Phase}, the model is tasked with sampling multiple candidate subgraphs (\textit{e.g.}, 5) while ensuring both their structural validity and factual existence. This constraint minimizes the risk of generating ``hallucinated'' or non-existent subgraphs. By enforcing diverse sampling criteria—ranging from a ``target-node-only'' structure to ``semantically-dense'' subgraphs—we guide the model to explore varying scales and semantic contexts. Furthermore, by optimizing for ``information difference'' between candidates, we encourage the model to consider distinct topological structures and textual attributes, thereby enhancing the overall quality and diversity of the subgraph pool.

In the \textbf{Select Phase}, the model is expected to autonomously evaluate the quality of candidate subgraphs and identify the optimal one. To facilitate this, we prescribe only fundamental constraints, granting the model maximum reasoning autonomy. This design allows the model’s internal selecting logic to evolve and refine adaptively during the training process, ensuring that the selection criteria align with the overall objective.

In the \textbf{Reason Phase}, we impose rigorous constraints by requiring the model to ``repeat and verify'' the selected subgraphs. This mechanism is designed to maintain logical consistency during long-context generation. By prioritizing independent analysis and holistic modeling, we effectively isolate the decision-making process from the interference of non-selected candidate subgraphs. This ensures that the final classification is derived strictly from the local topology and semantic evidence of the chosen subgraph, yielding the optimal prediction.

\subsection{Subgraph Diversity Score Calculation}\label{app:promot_diver}

\begin{table*}[htp!]
    \begin{tcolorbox}[
        title=\textbf{\textcolor{white}{A Prompt Sample for Subgraph Diversity Score Calculation}},
        colframe=black,
        coltitle=black,
        boxrule=1.0pt,
        arc=1mm,
        auto outer arc,
    ]
    \textbf{Task Description}:\\
    You are a graph analysis expert. Given two graph structures, which includes the description of the central node(s), its neighboring nodes, and their connection relationships, please follow the following steps to operate the analysis:\\
    
    \textbf{Graph structure 1}:\\
    - Central node ID(s): nodex\\
    - Node texts (central node(s) and neighboring nodes): [nodex: node description]\\
    - Connection relationships: [<nodex, nodey>]\\
    
    \textbf{Graph structure 2}:\\
    - Central node ID(s): nodex\\
    - Node texts (central node(s) and neighboring nodes): [nodex: node description]\\
    - Connection relationships: [<nodex, nodey>]\\
    
    \textbf{Step description}:\\
    1. Analyze the differences between the two graph structures in terms of the central node(s), neighboring nodes, and connection relationships.\\
    2. Based on the analysis, provide a distance score ranging from 0 to 1 to quantify how different these two graph structures are. A score of 0 indicates that the two graph structures are identical, while a score of 1 indicates that they are completely different.
    
    \end{tcolorbox}
\end{table*}

To capture both topological relationships and textual semantics across heterogeneous datasets, we move beyond traditional GNN-based embeddings. Instead, we employ LLMs to perform direct pairwise distance estimation. We task the LLM with performing a cross-graph structural and semantic analysis to generate a normalized score within the range of $[0, 1]$. This value serves as a universal distance metric to quantify the diversity between two subgraph structures. To further demonstrate the reliability of the LLM-based scoring, we conduct a stability analysis on 100 random samples across 100 independent trials each; the resulting average standard deviation for the Energy score was \textbf{0.05147} (relative to the metric's theoretical range of $(-\infty, -1]$), confirming that our LLM-based metric is robust and consistent for quantifying subgraph diversity.

\section{Training Details}

\paragraph{\textbf{Datasets}}

To ensure robust model performance across diverse zero-shot graph reasoning tasks, we developed specialized datasets for both SSR-SFT and SSR-RL. Our SSR-SFT training data originates from the publicly available Graph-R1 SFT collection. To adapt these samples to our specific task requirements, we utilized the SSR pipeline to reconstruct the corresponding prompts, which were then processed by DeepSeek-R1 to generate high-quality chain-of-thought reasoning paths and final answers. Following a rigorous quality filtration process, we curated a final SSR-SFT training set comprising 8,265 high-quality instances. Notably, in the Structural Diversity Evaluation, we set a predefined energy threshold of 0.3. For SSR-RL, we implemented a difficulty-aware sampling strategy to enhance the model's exploration capabilities. We first randomly sampled target nodes from the raw graph data and extracted their subgraphs. These subgraphs were transformed into prompts via the SSR pipeline and fed into the previously trained SSR-SFT model for an initial difficulty assessment. Specifically, each instance underwent five independent inference trials. An instance was categorized as ``Easy'' if the model produced correct results in 4 or 5 trials, ``Medium'' for 2 or 3 correct results, and ``Hard'' for 0 or 1 correct results. We sampled across these difficulty tiers with a fixed ratio of 2:2:1 (Easy:Medium:Hard), yielding a comprehensive SSR-RL training dataset of 10,000 instances. For evaluation, we adopt the datasets utilized in the GOFA paper. All validation instances were processed through the SSR pipeline to maintain consistency in prompt structure with the training phases, ensuring a reliable benchmark.

\paragraph{\textbf{Baselines}}

For the baseline results reported in Section~\ref{sec:q1}, we directly adopt the performance metrics provided in the Graph-R1 paper. To evaluate the Large Reasoning Models in Section~\ref{sec:q2}, we conducted evaluations on the same datasets using a prompt construction method without the SSR pipeline. Given that all experimental settings remain identical across these configurations, the results are directly comparable.

\paragraph{\textbf{Training Process}}\label{app:implement}

We adopt DeepSeek-R1-distilled-Qwen2.5-14B as the base model for all experiments to leverage its superior reasoning capabilities. The training process is divided into two distinct phases. First, in the SSR-SFT phase, we utilize the LlamaFactory framework to fine-tune the model for 3 epochs. We employ a learning rate of $1.0 \times 10^{-5}$ and a total batch size of 64 to establish a stable foundation for graph-based reasoning.

Subsequently, we transition to the SSR-RL phase using the verl framework. We adopt a sequential training strategy: Authenticity-Reinforced RLVR followed by Denoising-Reinforced RLVR. This phased approach is necessitated by the observation that the model’s graph reasoning capabilities following SSR-SFT are insufficient to manage the complexity of subgraph denoising. Introducing the reward $r_s$ prematurely could lead the model to aggressively prune subgraphs, losing vital information before it has learned how to reason on graph-based tasks. By first establishing a baseline of correctness through Authenticity-Reinforced RLVR, the model can then focus on refining its selections in Denoising-Reinforced RLVR. 

The overall optimization follows the GRPO loss formulation:
\begin{equation}
\begin{aligned}
    \mathcal{L}_{GRPO}(\theta) = & -\frac{1}{G} \sum_{i=1}^{G} \Biggl[ \min \Biggl( \frac{\pi_{\theta}(s_i|x)}{\pi_{\theta_{old}}(s_i|x)} \hat{A}_i, \\
    & \text{clip}\left(\frac{\pi_{\theta}(s_i|x)}{\pi_{\theta_{old}}(s_i|x)}, 1-\epsilon, 1+\epsilon\right) \hat{A}_i \Biggr) \\
    & - \beta \mathbb{D}_{KL}(\pi_{\theta} \| \pi_{ref}) \Biggr]
\end{aligned}
\end{equation}
where $G$ is the rollout group size and the advantage $\hat{A}_i$ is computed by normalizing the rewards within each group, driving the model to prefer outputs that achieve higher accuracy with lower structural noise.

For RL training, the model is trained for 4 epochs with a more conservative learning rate of $1.0 \times 10^{-6}$ and a batch size of 64. To prevent the model from being prematurely influenced by the denoising reward—especially when its fundamental graph reasoning capabilities are still maturing—we introduce Denoising-Reinforced RLVR solely during the final 2,048 instances of the RL training process. For the Denoising-Reinforced RLVR, we specifically set the denoising intensity hyperparameter $\lambda$ to 0.1, balancing the correction of noisy graph structures with the preservation of essential topological information. The entire pipeline, including data synthesis and model training, was executed on a high-performance cluster equipped with 8 NVIDIA H100 GPUs.

\section{Dataset Descriptions}\label{app:datasets}

\begin{table*}[htp!]
\centering
\caption{Comprehensive statistics of the multi-domain benchmarks. We employ a zero-shot setting, ensuring that datasets marked for testing remain entirely inaccessible during the training phase to rigorously assess model generalizability.}
\label{tab:datasets}
\small
\begin{tabular}{llcccc}
\toprule
\textbf{Domain} & \textbf{Dataset} & \textbf{\#Nodes} & \textbf{\#Edges} & \textbf{\#Classes} & \textbf{Split} \\ \midrule
\textbf{Social Network} & Instagram & 11,339 & 155,349 & 2 & Training \\ \midrule
\multirow{2}{*}{\textbf{Knowledge Graph}} & WN18RR & 40,943 & 93,003 & 11 & Training \\
 & \textbf{FB15K237} & \textbf{14,541} & \textbf{310,116} & \textbf{237} & \textbf{Test (Zero-shot)} \\ \midrule
\multirow{4}{*}{\textbf{Citation}} & Arxiv & 169,343 & 1,166,243 & 40 & Training \\
 & Citeseer & 3,186 & 8,554 & 6 & Training \\
 & Pubmed & 19,717 & 88,648 & 3 & Training \\
 & \textbf{Cora} & \textbf{2,708} & \textbf{10,556} & \textbf{7} & \textbf{Test (Zero-shot)} \\ \midrule
\multirow{5}{*}{\textbf{E-commerce}} & Children & 76,875 & 1,554,578 & 24 & Training \\
 & Computer & 87,229 & 721,081 & 10 & Training \\
 & Photo & 48,362 & 500,939 & 12 & Training \\
 & Sports & 173,055 & 1,773,500 & 13 & Training \\
 & \textbf{Products} & \textbf{54,025} & \textbf{144,638} & \textbf{47} & \textbf{Test (Zero-shot)} \\ \midrule
\textbf{Web Link} & \textbf{WikiCS} & \textbf{11,701} & \textbf{216,123} & \textbf{10} & \textbf{Test (Zero-shot)} \\ \bottomrule
\end{tabular}
\end{table*}

To verify the cross-dataset adaptability and zero-shot performance of GraphSSR, we conducted experiments on several benchmark datasets spanning five distinct domains: Social Networks, Knowledge Graphs, Citation Networks, E-commerce, and Web Links. Detailed topological statistics for these datasets are summarized in Table~\ref{tab:datasets}.

\paragraph{\textbf{Social Network}}

\begin{itemize}
    \item \textbf{Instagram}~\cite{li2024glbench}: This dataset represents a social graph where nodes signify individual users and edges illustrate social ties (e.g., follower-following dynamics). Node-level features are derived from user-provided profile summaries and self-descriptions.
\end{itemize}

\paragraph{\textbf{Knowledge Graph}}

\begin{itemize}
    \item \textbf{WN18RR}~\cite{liu2023one}: A subset of the WordNet lexical database, this graph maps relationships between English words. It consists of 40,943 nodes where edges define semantic connections between linguistic entities.
    \item \textbf{FB15K237}~\cite{liu2023one}: Derived from Freebase, this large-scale knowledge base contains entities (nodes) and their multifaceted relations (edges). Each entity is enriched with textual descriptions based on its name and relational context.
\end{itemize}

\paragraph{\textbf{Citation}}

\begin{itemize}
    \item \textbf{Arxiv}~\cite{hu2020open}: A massive citation network of computer science papers. Each node is an academic publication, and edges denote direct citations. The objective is to categorize papers into 40 distinct subfields.
    \item \textbf{Citeseer}~\cite{yang2016revisiting}: A specialized citation network centered on the computer science domain. It maps the interconnectedness of research papers through their citation links.
    \item \textbf{Pubmed}~\cite{he2023harnessing}: A biomedical citation graph sourced from the PubMed database. Nodes represent scientific articles related to diseases, and edges represent citation links.
    \item \textbf{Cora}~\cite{wen2023augmenting}: A machine learning-centric citation graph. Despite its smaller size, it provides a challenging classification task involving 7 fine-grained categories of research papers.
\end{itemize}

\paragraph{\textbf{E-commerce}}

\begin{itemize}
    \item \textbf{Children, Computer, Photo, Sports, \& Products}~\cite{yan2023comprehensive, he2023harnessing}: These datasets are product graphs from Amazon. Nodes represent specific products, and edges indicate that products were frequently viewed or purchased in the same session. Textual metadata includes user reviews.
\end{itemize}

\paragraph{\textbf{Web Link}}

\begin{itemize}
    \item \textbf{WikiCS}~\cite{mernyei2020wiki}: A hyperlink network focused on the computer science branch of Wikipedia. Nodes correspond to individual articles, and directed edges represent the web links connecting them. The full text of each article provides the rich semantic information for each node.
\end{itemize}

\section{Computational Cost and Inference Efficiency Analysis}\label{app:computational_cost}

To provide a comprehensive overview of the computational demands of GraphSSR, we systematically evaluate its inference efficiency across all datasets. As presented in Table~\ref{tab:inference_efficiency}, GraphSSR consumes 6,981.7 tokens with a latency of 208.2s on average, compared to 1,098.4 tokens and 27.6s for Graph-R1.

To investigate the root cause of this variance, we provide a granular step-by-step cost breakdown for GraphSSR in Table~\ref{tab:cost_breakdown}. The analysis reveals that the computational overhead is heavily concentrated, with over 70\% of the resource consumption dedicated to the Sample Phase (57.26\%) and answer standardization (14.20\%). 

Although GraphSSR entails increased inference costs, we argue that this is a necessary and justified trade-off for its capacity to autonomously explore the complex subgraph space and successfully mitigate detrimental structural noise. To further improve practical feasibility, we are also exploring a distillation strategy where GraphSSR serves as a teacher. Specifically, the teacher provides high-quality denoised subgraph examples and accurate reasoning traces to help a student model learn to directly generate an optimized subgraph in a single step. By merging the Sample and Select phases into a unified generative process, we can reduce token consumption and latency while retaining the teacher's denoising expertise. We leave this for future work to balance state-of-the-art performance with computational efficiency.

\begin{table}[h]
    \centering
    \caption{Inference Efficiency Comparison}
    \label{tab:inference_efficiency}
    \begin{tabular}{lccc}
    \toprule
    \textbf{Metric} & \textbf{Graph-R1} & \textbf{GraphSSR} & \textbf{Diff (\%)} \\
    \midrule
    Tokens & 1,098.4 & 6,981.7 & +535.62\% \\
    Latency (s) & 27.6 & 208.2 & +654.13\% \\
    \bottomrule
    \end{tabular}
\end{table}

\begin{table*}[t]
    \centering
    \caption{Breakdown of GraphSSR Computation Cost}
    \label{tab:cost_breakdown}
    \begin{tabular}{lcccc}
    \toprule
    \textbf{Metric} & \textbf{Sample Phase} & \textbf{Select Phase} & \textbf{Reason Phase} & \textbf{Answer Std.} \\
    \midrule
    Tokens & 3,262 & 495 & 1,129 & 809 \\
    Ratio (\%) & 57.26\% & 8.70\% & 19.83\% & 14.20\% \\
    \bottomrule
    \end{tabular}
\end{table*}

\section{Detailed Analysis of Case Study on Cora}

To further elucidate the internal mechanisms of the Sample-Select-Reason (SSR) pipeline and the impact of the two-stage RL on subgraph denoising, we present a comprehensive breakdown of the case study referenced in the main text. This example from the Cora dataset illustrates how structural noise in a dense egocentric subgraph can interfere with graph reasoning and how our strategy successfully recovers the precise semantic signal.

\subsection{Input Data Characteristics and Semantic Dissonance}

The target node (\textbf{node 11}) is defined by the text: ``\textit{Hierarchical logistic belief networks for linear model selection, using Gibbs sampling for parameter learning}''. The ground-truth label is \textbf{``Neural Networks''}.

Due to spatial constraints, the ``Input Egocentric Subgraph'' (\textit{i.e.}, the uniform subgraph used in Graph-R1) exclusively visualizes the target node's first-order neighbors, omitting second-order and higher-order neighbors. These neighboring nodes may provide task-relevant signals, they can also introduce potential noise. Specifically,

\begin{itemize}
    \item \textbf{Relevant Nodes:} \textbf{node 13} (hierarchical mixtures of experts) and \textbf{node 17} (representations in neural networks) are all nodes related to the network architecture. They are all directly connected to the target node, forming a ``Network'' semantic cluster.
    \item \textbf{Noisy Nodes:} \textbf{node 9} (EM algorithm) and \textbf{node 14} (EM reinterpreted via free energy maximization) describe the probabilistic method about the EM algorithm. While topically related, their semantic relevance to the target node is not strong enough.
\end{itemize}

\definecolor{modelbg}{RGB}{153, 76, 0}
\definecolor{modelframe}{RGB}{153, 76, 0}

\begin{table*}[b]
    \begin{tcolorbox}[
        title=\textbf{\textcolor{white}{The Input Egocentric Subgraph}},
        colframe=modelbg,
        coltitle=modelframe,
        boxrule=1.0pt,
        arc=1mm,
        auto outer arc,
    ]
    \textbf{Complete subgraph:}\\
    - \textbf{Central node ID(s):} node11\\
    - \textbf{Node texts (central node(s) and neighboring nodes):} [\\
    \hspace*{1em}node9: EM algorithm for mixtures of factor analyzers, combining clustering and dimensionality reduction in latent variable models.\\
    \hspace*{1em}node11: Hierarchical logistic belief networks for linear model selection, using Gibbs sampling for parameter learning.\\
    \hspace*{1em}node13: Tree-structured hierarchical mixtures of experts with EM algorithm, validated via robot dynamics simulations.\\
    \hspace*{1em}node14: EM reinterpreted via free energy maximization, enabling faster incremental updates and improved convergence.\\
    \hspace*{1em}node17: Generative models for sparse distributed representations in neural networks, using Bayesian inference and simple learning rules.\\
    ]\\
    - \textbf{Connection relationships:} [\\
    \hspace*{1em}<node0, node8>, <node1, node12>, <node1, node14>, <node2, node13>, <node3, node13>, <node3, node14>, <node3, node20>, <node4, node13>, <node4, node18>, <node5, node13>, <node6, node7>, <node7, node13>, <node8, node17>, <node9, node11>, <node10, node13>, <node10, node14>, <node10, node18>, <node11, node13>, <node11, node14>, <node11, node17>, <node13, node18>, <node13, node21>, <node15, node18>, <node16, node18>, <node18, node19>\\
    ]
    \end{tcolorbox}
\end{table*}

\subsection{The Necessity of SSR Pipeline in Task-Relevant Subgraph Denoising}\label{app:case_ssr}

In this section, we provide an extended comparative analysis of the real-world inference results for the case study introduced in Section~\ref{exp:case_qualitative}, contrasting GraphSSR with its variant that omits the SSR pipeline. This comparison serves to further justify the necessity of the SSR pipeline and the specialized training regime in isolating task-relevant structural information from inherent graph noise. The detailed variants and results are presented below:

\paragraph{\textbf{Variant: w/o SSR pipeline.}} In this variant, the model is forced to process all neighbors simultaneously, including all first-order neighboring nodes and the unlisted second-order neighboring nodes, etc. This results in the model’s reasoning focuses on the cumulative occurrence of keywords like ``EM algorithm''. The signals from the nodes describing ``networks''  are drowned out by the statistical prevalence of EM-related neighbors. Consequently, the model fails to capture the ``Neural Networks'' essence of the target node and yields a biased classification of ``Probabilistic Methods''.

\paragraph{\textbf{GraphSSR.}} The full model GraphSSR, equipped with the SSR pipeline and trained with the two-stage RL, demonstrates a sophisticated filtering strategy. It samples five candidate subgraphs ($Subgraph\_0$ to $Subgraph\_4$), including a pure subgraph $Subgraph\_2$ which contains only node 13 and node 17, and a subgraph $Subgraph\_4$ which contains all the first-order neighbors. Due to the existence of the subgraph denoising mechanism, GraphSSR selects $Subgraph\_2$ rather than the larger $Subgraph\_4$, and ``courageously'' prunes the noisy nodes describing EM algorithms. By isolating the relationship between ``belief networks'', ``mixtures of experts'', and ``neural networks'', the LLM's reasoning engine recognizes a unified theme: network architecture. This distilled context allows it to bypass the ``Probabilistic Methods'' category and correctly identify the specific ``Neural Networks'' category.

\begin{table*}[b]
    \begin{tcolorbox}[
        title=\textbf{\textcolor{white}{The Output of the ``w/o SSR Pipeline'' Variant}},
        colframe=modelbg,
        coltitle=modelframe,
        boxrule=1.0pt,
        arc=1mm,
        auto outer arc,
    ]
    \textbf{Answer}: Probabilistic Methods  \\
    \textbf{Brief\_reasoning}: Node11 focuses on ``hierarchical logistic belief networks'' and Gibbs sampling, aligning with probabilistic modeling techniques. Connections to nodes involving EM algorithms (node9, node10, node14), mixture models (node13), and Gibbs sampling (node15, node18) reinforce its probabilistic methodology. While ``belief networks'' can intersect with neural networks, the emphasis on hierarchical structure and sampling for parameter learning fits better under probabilistic methods. Neural Networks is less precise here, as no direct mention of neural architecture (e.g., layers, activation functions) exists.
    \end{tcolorbox}
\end{table*}

\begin{table*}[htp!]
    \begin{tcolorbox}[
        title=\textbf{\textcolor{white}{The Output of GraphSSR}},
        colframe=modelbg,
        coltitle=modelframe,
        boxrule=1.0pt,
        arc=1mm,
        auto outer arc,
    ]
    \textbf{Subgraph\_0}\\
    \hspace*{1em}Central\_node\_ID: node11\\
    \hspace*{1em}Neighboring\_node\_ID: \\
    \hspace*{1em}Connection\_relationship: \\
    
    \textbf{Subgraph\_1}\\
    \hspace*{1em}Central\_node\_ID: node11\\
    \hspace*{1em}Neighboring\_node\_ID: node13\\
    \hspace*{1em}Connection\_relationship: <node11, node13>\\
    
    \textbf{Subgraph\_2}\\
    \hspace*{1em}Central\_node\_ID: node11\\
    \hspace*{1em}Neighboring\_node\_ID: node13,node17\\
    \hspace*{1em}Connection\_relationship: <node11, node13>,<node11, node17>\\
    
    \textbf{Subgraph\_3}\\
    \hspace*{1em}Central\_node\_ID: node11\\
    \hspace*{1em}Neighboring\_node\_ID: node9,node13,node14,node17\\
    \hspace*{1em}Connection\_relationship: <node9, node11>,<node11, node13>,<node11, node14>,<node11, node17>\\
    
    \textbf{Subgraph\_4}\\
    \hspace*{1em}Central\_node\_ID: node11\\
    \hspace*{1em}Neighboring\_node\_ID: node13,node4,node18\\
    \hspace*{1em}Connection\_relationship: <node11, node13>,<node4, node13>,<node4, node18>,<node13, node18>\\
    
    \textbf{Chosen\_subgraph\_reasoning}\\
    \hspace*{1em}\textbf{Chosen\_subgraph:} 2\\
    \hspace*{1em}Chosen\_subgraph\_reason: This subgraph optimally balances relevance and focus: node13 directly supports hierarchical modeling, while node17 explicitly references neural networks, providing strong semantic reinforcement for classification without extraneous nodes.\\
    \hspace*{1em}Central\_node\_ID: node11\\
    \hspace*{1em}Neighboring\_node\_ID: node13,node17\\
    \hspace*{1em}Connection\_relationship: <node11, node13>,<node11, node17>\\
    \hspace*{1em}\textbf{Answer:} Neural Networks\\
    \hspace*{1em}\textbf{Brief\_reasoning:} Node11 describes ``hierarchical logistic belief networks'' (a neural network architecture). Node13 discusses ``hierarchical mixtures of experts'' (a neural network model), and node17 explicitly mentions ``neural networks'' in generative modeling. Collective context confirms neural network focus.
    \end{tcolorbox}
\end{table*}

\subsection{The Synergistic Effect of Two-Stage RL in Balancing Denoising and Reasoning}\label{app:case_rl}

As illustrated in Section~\ref{sec:method_rl}, we introduce SSR-RL, a novel two-stage reinforcement learning framework, to enhance the model's capacity for both subgraph denoising and complex reasoning. To further demonstrate how this two-stage fusion approach can enhance the ability of subgraph denoising, we conduct a comparative analysis using two distinct model variants: one employing solely Authenticity-Reinforced RLVR and the other restricted to Denoising-Reinforced RLVR.

\paragraph{\textbf{Variant: solely Authenticity-Reinforced RLVR.}} Based on the SSR pipeline, the variant can also sample five candidate subgraphs and include a relatively pure subgraph $Subgraph\_3$ in the candidates. However, without the subgraph denoising reward from Denoising-Reinforced RLVR, it incorrectly assumes that more data equals better evidence and chooses $Subgraph\_4$, which is the most comprehensive (containing all 4 first-order neighbors). Since $Subgraph\_4$ still contains noisy node 9 and node 14, the reasoning is flawed and leads to the incorrect answer ``Probabilistic Methods''.

\paragraph{\textbf{Variant: solely Denoising-Reinforced RLVR.}} In this variant, there is also a relatively pure subgraph $Subgraph\_1$ with only the neighboring node 13 in the candidates. However without Authenticity-Reinforced RLVR, the model has not yet developed a strong graph reasoning ability at the early stage of training. The premature introduction of the subgraph denoising reward would lead the model to favor smaller, information-sparse subgraphs. Specifically, the model selects $Subgraph\_0$—containing only the isolated target node itself—to avoid potential neighbor noise, yet this strategy backfires. Without the structural context of neighboring nodes to clarify its identity, the model is misled by the ``Gibbs sampling'' description in the target node 11 and also incorrectly yields a ``Probabilistic Methods'' label.

The empirical results from these two variants demonstrate that Authenticity-Reinforced RLVR and Denoising-Reinforced RLVR serve as indispensable components of GraphSSR. Specifically, the absence of either module results in an imbalance in the adaptive subgraph denoising process—either failing to filter out sufficient noise or inadvertently removing critical structural information. Such an imbalance inevitably leads to suboptimal subgraph quality and a subsequent degradation in overall model performance. By synergistically integrating these two modules, GraphSSR successfully strikes a robust balance between noise reduction and information preservation, ensuring that the generated subgraphs remain both pure and informative, ultimately driving the superior performance of the model across various benchmarks.

\begin{table*}[htp!]
    \begin{tcolorbox}[
        title=\textbf{\textcolor{white}{The Output of the ``solely Authenticity-Reinforced RLVR'' Variant}},
        colframe=modelbg,
        coltitle=modelframe,
        boxrule=1.0pt,
        arc=1mm,
        auto outer arc,
    ]
    \textbf{Subgraph\_0}\\
    \hspace*{1em}Central\_node\_ID: node11\\
    \hspace*{1em}Neighboring\_node\_ID: \\
    \hspace*{1em}Connection\_relationship: \\
    
    \textbf{Subgraph\_1}\\
    \hspace*{1em}Central\_node\_ID: node11\\
    \hspace*{1em}Neighboring\_node\_ID: node9\\
    \hspace*{1em}Connection\_relationship: <node9, node11>\\
    
    \textbf{Subgraph\_2}\\
    \hspace*{1em}Central\_node\_ID: node11\\
    \hspace*{1em}Neighboring\_node\_ID: node13,node14\\
    \hspace*{1em}Connection\_relationship: <node11, node13>,<node11, node14>\\
    
    \textbf{Subgraph\_3}\\
    \hspace*{1em}Central\_node\_ID: node11\\
    \hspace*{1em}Neighboring\_node\_ID: node13,node18\\
    \hspace*{1em}Connection\_relationship: <node11, node13>,<node13, node18>\\
    
    \textbf{Subgraph\_4}\\
    \hspace*{1em}Central\_node\_ID: node11\\
    \hspace*{1em}Neighboring\_node\_ID: node9,node13,node14,node17\\
    \hspace*{1em}Connection\_relationship: <node9, node11>,<node11, node13>,<node11, node14>,<node11, node17>\\
    
    \textbf{Chosen\_subgraph\_reasoning}\\
    \hspace*{1em}\textbf{Chosen\_subgraph:} 4\\
    \hspace*{1em}Chosen\_subgraph\_reason: This subgraph contains all direct neighbors of node11, providing maximal semantic context about probabilistic methods (EM, Gibbs sampling) and neural networks, which are central to node11's description.\\
    \hspace*{1em}Central\_node\_ID: node11\\
    \hspace*{1em}Neighboring\_node\_ID: node9,node13,node14,node17\\
    \hspace*{1em}Connection\_relationship: <node9, node11>,<node11, node13>,<node11, node14>,<node11, node17>\\
    \hspace*{1em}\textbf{Answer:} Probabilistic Methods\\
    \hspace*{1em}\textbf{Brief\_reasoning:} The central node (node11) describes hierarchical logistic belief networks using Gibbs sampling. Neighbors node9, node13, and node14 focus on EM algorithms for probabilistic models, while node17 discusses generative models with Bayesian inference. Collectively, this subgraph emphasizes probabilistic modeling techniques, aligning with the ``Probabilistic Methods'' classification.
    \end{tcolorbox}
\end{table*}

\begin{table*}[htp!]
    \begin{tcolorbox}[
        title=\textbf{\textcolor{white}{The Output of the ``solely Denoising-Reinforced RLVR'' Variant}},
        colframe=modelbg,
        coltitle=modelframe,
        boxrule=1.0pt,
        arc=1mm,
        auto outer arc,
    ]
    \textbf{Subgraph\_0}\\
    \hspace*{1em}Central\_node\_ID: node11\\
    \hspace*{1em}Neighboring\_node\_ID: \\
    \hspace*{1em}Connection\_relationship: \\
    
    \textbf{Subgraph\_1}\\
    \hspace*{1em}Central\_node\_ID: node11\\
    \hspace*{1em}Neighboring\_node\_ID: node13\\
    \hspace*{1em}Connection\_relationship: <node11, node13>\\
    
    \textbf{Subgraph\_2}\\
    \hspace*{1em}Central\_node\_ID: node11\\
    \hspace*{1em}Neighboring\_node\_ID: node13,node14\\
    \hspace*{1em}Connection\_relationship: <node11, node13>,<node11, node14>\\
    
    \textbf{Subgraph\_3}\\
    \hspace*{1em}Central\_node\_ID: node11\\
    \hspace*{1em}Neighboring\_node\_ID: node17,node8\\
    \hspace*{1em}Connection\_relationship: <node11, node17>,<node8, node17>\\
    
    \textbf{Subgraph\_4}\\
    \hspace*{1em}Central\_node\_ID: node11\\
    \hspace*{1em}Neighboring\_node\_ID: node9,node13,node14,node17,node18\\
    \hspace*{1em}Connection\_relationship: <node9, node11>,<node11, node13>,<node11, node14>,<node11, node17>,<node13, node18>\\
    
    \textbf{Chosen\_subgraph\_reasoning}\\
    \hspace*{1em}\textbf{Chosen\_subgraph:} 0\\
    \hspace*{1em}Chosen\_subgraph\_reason: Subgraph\_0 is optimal as it isolates the central node's core semantics (``belief networks'' and ``Gibbs sampling''), avoiding potential neighbor-induced biases (e.g., neural network terminology in other subgraphs). The central node's text alone provides unambiguous evidence for classification.\\
    \hspace*{1em}Central\_node\_ID: node11\\
    \hspace*{1em}Neighboring\_node\_ID: \\
    \hspace*{1em}Connection\_relationship: \\
    \hspace*{1em}\textbf{Answer:} Probabilistic Methods\\
    \hspace*{1em}\textbf{Brief\_reasoning:} The central node (node11) explicitly describes hierarchical logistic belief networks—a probabilistic graphical model, and uses Gibbs sampling—a core probabilistic inference technique. This directly aligns with ``Probabilistic Methods'' without requiring neighbor context.
    \end{tcolorbox}
\end{table*}

\end{document}